\documentclass{article} % For LaTeX2e
\usepackage{iclr2024_conference,times}

\usepackage{amsmath,amsfonts,bm}

\def\eqref#1{equation~\ref{#1}}
\def\1{\bm{1}}

\DeclareMathAlphabet{\mathsfit}{\encodingdefault}{\sfdefault}{m}{sl}
\SetMathAlphabet{\mathsfit}{bold}{\encodingdefault}{\sfdefault}{bx}{n}

\usepackage{hyperref}
\usepackage{url}
\usepackage{booktabs}       % professional-quality tables
\usepackage{graphicx}
\usepackage{amsmath}
\usepackage{multirow}
\usepackage{multicol}
\usepackage{wrapfig}
\usepackage{amssymb}% http://ctan.org/pkg/amssymb
\usepackage{pifont}% http://ctan.org/pkg/pifont
\newcommand{\cmark}{\ding{51}}%

\usepackage[capitalize]{cleveref}
\crefname{section}{Sec.}{Secs.}
\Crefname{section}{Section}{Sections}
\Crefname{table}{Table}{Tables}
\crefname{table}{Tab.}{Tabs.}

\newcommand{\et}[2]{${#1}^{\pm{#2}}$}
\newcommand{\etb}[2]{$\mathbf{{#1}}^{\pm{#2}}$}
\newcommand{\ets}[2]{$\underline{{#1}}^{\pm{#2}}$}

\newcommand{\revise}[1]{#1}

\title{Generative Human Motion Stylization in \\Latent Space}

\author{Chuan Guo\textsuperscript{1*},\quad Yuxuan Mu\textsuperscript{1*},\quad Xinxin Zuo\textsuperscript{2},\quad Peng Dai\textsuperscript{2},\quad Youliang Yan\textsuperscript{2},\\
\textbf{Juwei Lu\textsuperscript{2}, \quad Li Cheng\textsuperscript{1}}  \\
\textsuperscript{1}University of Alberta, \quad \textsuperscript{2}Noah’s Ark Lab, Huawei Canada
}

\iclrfinalcopy % Uncomment for camera-ready version, but NOT for submission.
\begin{document}

\def\thefootnote{*}\footnotetext{These authors contributed equally to this work.}\def\thefootnote{\arabic{footnote}}

\maketitle

\begin{abstract}
Human motion stylization aims to revise the style of an input motion while keeping its content unaltered. Unlike existing works that operate directly in pose space, we leverage the \textit{latent space} of pretrained autoencoders as a more expressive and robust representation for motion extraction and infusion. Building upon this, we present a novel \textit{generative} model that produces diverse stylization results of a single motion (latent) code. During training, a motion code is decomposed into two coding components: a deterministic content code, and a probabilistic style code adhering to a prior distribution; then a generator massages the random combination of content and style codes to reconstruct the corresponding motion codes. Our approach is versatile, allowing the learning of probabilistic style space from either style labeled or unlabeled motions, providing notable flexibility in stylization as well. In inference, users can opt to stylize a motion using style cues from a reference motion or a label. Even in the absence of explicit style input, our model facilitates novel re-stylization by sampling from the unconditional style prior distribution. Experimental results show that our proposed stylization models, despite their lightweight design, outperform the state-of-the-arts in style reeanactment, content preservation, and generalization across various applications and settings.
%Human motion stylization aims to update the style of an input motion while keeping its content unaltered. Unlike previous works that are usually limited to deterministic outcomes, we propose a \textit{generative} framework that produces diverse stylization results given an 3D human motion and style signal. Our key insight is to decompose a motion sequence into two independent coding components: a temporal content code and a probabilistic style space confined by a fixed prior Gaussian distribution; this style space can be conditioned on an additional style label. Then a generator projects proper combinations of content and style back to motion. During inference, the style input can be obtained either from a reference motion (\textbf{motion-based}) or by sampling directly from the prior distribution. In the latter scenario, the conditional style space allows for \textbf{label-based} diverse stylization; while in unconditional setting, sampling from the prior distribution results in \textbf{prior-based} novel motion re-stylization. Moreover, instead of directly stylizing 3D poses, we propose \textit{latent stylization}, where style extraction and injection are applied to motion latent features, a more expressive and compact motion representation from pre-trained autoencoders. Experiments on three human motion benchmarks demonstrate the superior performance of our approach on style reenactment, content preservation and generalization capability.
\end{abstract}

\section{Introduction}
%Our movements are much more expressive than we can imagine, which could convey information about our personality, mood and even occupation. For example, we could often identify a person, or tell the mood of a person, from the way they walk. These distinguishable motion traits, usually thought of as styles, have been essential elements in 3D industry for realistic character animation. However, it is expensive to capture motions with various styles. Stylizing existing motions using a reference style motion (\textit{i.e.,} motion-based), or a preset style label (\textit{i.e.,} label-based) becomes a feasible solution. 
The motions of our humans are very expressive and contain a rich source of information. For example, by watching a short duration of one individual's walking movement, we could quickly recognize the person, or discern the mood, age, or occupation of the person. These distinguishable motion traits, usually thought of as styles, are therefore essential in film or game industry for realistic character animation. It is unfortunately unrealistic to acquire real-world human motions of various styles solely by motion capture. Stylizing existing motions using a reference style motion (\textit{i.e.,} motion-based), or a preset style label (\textit{i.e.,} label-based) thus becomes a feasible solution. 

%Existing works model the stylization process in either deterministic~\citep{aberman2020unpaired,holden2016deep,jang2022motion,tao_style-erd_2022} or non-deterministic ~\citep{park2021diverse}
Deep learning models have recently enabled numerous data-driven methods for human motion stylization. These approaches, however, still find their shortfalls. A long line of existing works~\citep{aberman2020unpaired,holden2016deep,jang2022motion,tao_style-erd_2022} are limited to deterministic stylization outcomes. ~\citep{park2021diverse,wen2021autoregressive} though allows diverse stylization, their results are far from being satisfactory, and the trained models struggle to generalize to other motion datasets. Furthermore, all of these approaches directly manipulate style within raw poses, a redundant and potentially noisy representation of motions. Meanwhile, they often possess rigid designs, allowing for only supervised or unsupervised training, with style input typically limited to either reference motions or labels, as shown in~\cref{tab:flexibility}.

In this work, we introduce a novel \textit{generative} stylization framework for 3D human motions. Inspired by the recent success of content synthesis in latent space~\citep{guo2022generating, chen2022executing,rombach2022high,ramesh2022hierarchical}, we propose to use \textit{latent} motion features (namely motion code) of pretrained convolutional autoencoders as the intermedia for motion style extraction and infusion. Compared to raw poses, the benefits are three-folds: \textbf{(i)} Motion codes are more compact and expressive, containing the most discriminative features of raw motions; \textbf{(ii)} Autoencoders can be learned once on a large dataset and reused for downstream datasets. Thanks to the inductive bias of CNN~\citep{lecun1995convolutional}, the learned motion code features typically contains less noise, resulting in improved generalization, as empirically demonstrated in~\Cref{tab:aberman_cmu,tab:aberman_xia}; \textbf{(iii)} Practically, motion code sequences are much shorter than motions, making them more manageable in neural networks. Building on this, our latent stylization framework decomposes the motion code into two components: a temporal and deterministic \textit{content} code, and a global probabilistic \textit{style} code confined by a prior Gaussian distribution. The subsequent generator recombines content and style to synthesize valid motion code. During training, besides auto-encoding and decoding, we swap the contents and styles between random pairs, and the resulting motion codes are enforced to recover the source contents and styles through cycle reconstruction. To further improve content-style disentanglement, we propose a technique called \textit{homo-style alignment}, which encourages the alignment of style spaces formed by different motion sub-clips from the same sequence. Lastly, the global velocity of resulting motions are obtained through a pre-trained global motion regressor.

\begin{figure*}[t]
	\centering
	\includegraphics[width=0.85\linewidth]{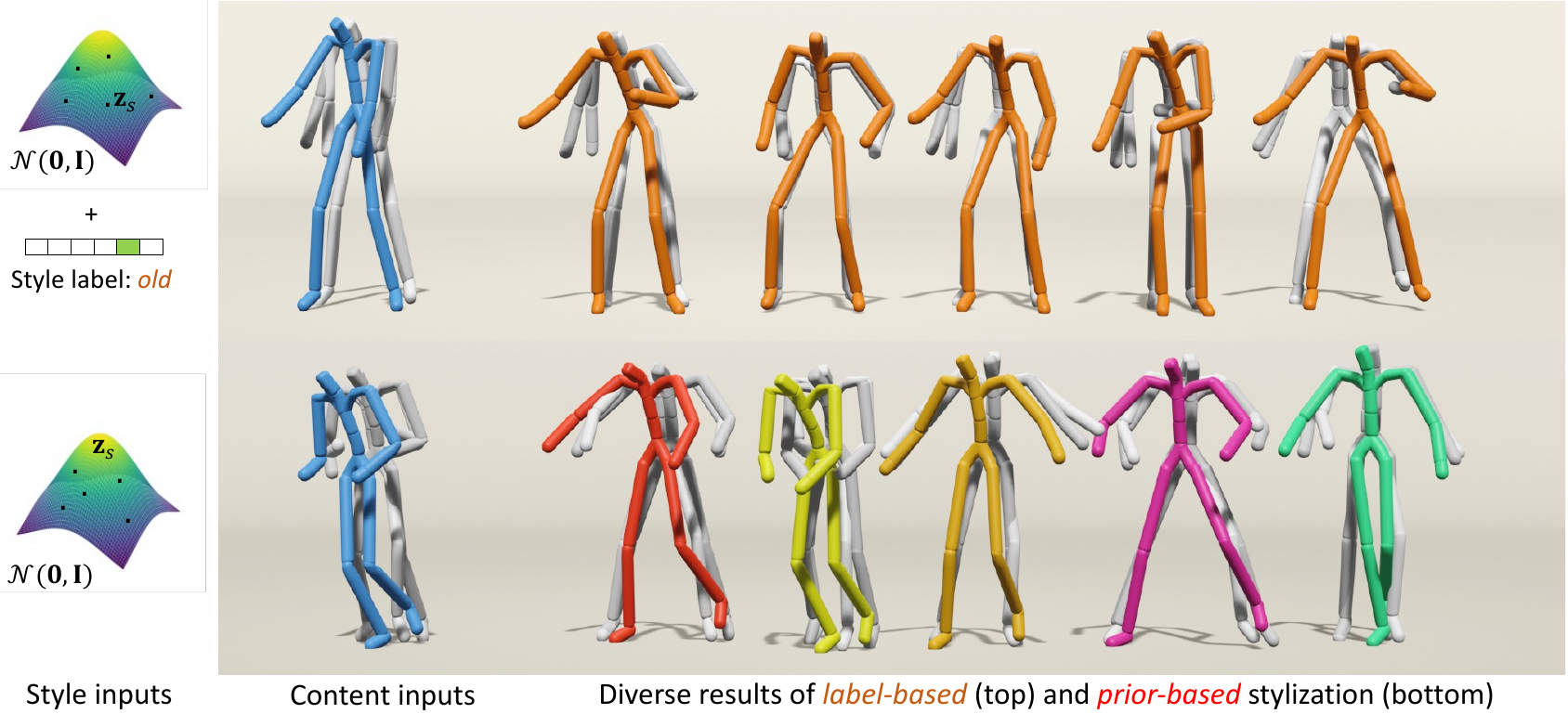}
 	 \vspace{-0.5em}
	\caption{\footnotesize{(\textbf{Top}) Given an input motion and target style label (\textit{i.e., old}), our label-based stylization generates diverse results following provided label. (\textbf{Bottom}) Without any style indicators, our prior-based method randomly re-stylizes the input motion using sampled prior styles $\mathbf{z}_s$. Five distinct stylized motions from the same content are presented, with poses synchronized and history in gray. See~\cref{fig:inference} (b) and (d) for implementations.}}
	\label{fig:teaser}
	 \vspace{-1.8em}
\end{figure*}

Our approach offers versatile stylization capabilities (\cref{tab:flexibility}), accommodating various conditioning options during both training and inference: 1) Deterministic stylization using style from \textbf{exemplar motions}; 2) In the label conditioned setting, our model can perform diverse stylization based on provided \textbf{style labels}, as in ~\cref{fig:teaser} (top); 3) In the unconditional setting, our model can randomly sample styles from the \textbf{prior distribution} to achieve stochastic stylization, as in~\cref{fig:teaser} (bottom). Benefiting from our latent stylization and lightweight model design, our approach achieves state-of-the-art performance while being 14 times faster than the most advanced prior work~\citep{jang2022motion}, as shown in \Cref{tab:infertime}.

Our key contributions can be summarized as follows. Firstly, we propose a novel generative framework, using motion latent features as an advanced alternative representation, accommodating various training and inference schemes in a single framework. Secondly, through a comprehensive suite of evaluations on three benchmarks, our framework demonstrates robust and superior performance across all training and inference settings, with notable efficiency gains.
\section{Related Work}

\paragraph{Image Style Transfer.} Image style in computer vision and graphics is typically formulated as the global statistic features of images. Early work~\citep{gatys2016image} finds it possible to transfer the visual style from one image to another through aligning their Gram matrices in neural networks. On top of this, ~\citep{johnson2016perceptual,ulyanov2016texture} enable faster transferring through an additional feed-forward neural networks. The work of ~\citep{ulyanov2016instance} realizes that instance normalization (IN) layer could lead to better performance. However, these works can only be applied on single style image. 
%The following efforts~\citep{chen2017stylebank,zhang2018multi,dumoulin2016learned} work on one-model multi-style transfer which maintains a candidate pools of style parameters for pre-defined set of image styles. 
~\citep{huang2017arbitrary} facilitates arbitrary image style transfer by introducing adaptive instance normalization (AdaIN).
%Specifically, AdaIN modifies the mean and variance of the deep feature maps of content image using learned affine parameters from style image. The effectiveness of IN and AdaIN editing style information has also been validated in several works such as CycleGAN~\citep{zhu2017unpaired} and styleGAN~\citep{karras2019style}. The aforementioned parametric image style transfer methods usually fail in detailed structure synthesis. 
Alternatively, in PatchGAN~\citep{isola2017image} and CycleGAN~\citep{zhu2017unpaired}, textures and styles are translated between images by ensuring the local similarity using patch discriminator. Similar idea was adopted in ~\citep{park2020swapping}, which proposes patch co-occurrence discriminator that hypothesizes images with similar marginal and joint feature statistics appear perceptually similar.

\begin{table}[t]
    \centering
    \scalebox{0.7}{
    \begin{tabular}{l c c c c c c c}
    \toprule
      ~& \multicolumn{2}{c}{Supervised (\textit{w} style label)} & ~ &\multicolumn{2}{c}{Unsupervised (\textit{w/o} style label)}& ~ &\multirow{2}{*}{Generative}\\
      \cline{2-3}
      \cline{5-6}
      ~& Motion-based & Label-based & ~& Motion-based & Prior-based&  \\
         \midrule
         % \midrule
         \citep{xia2015realtime}& ~ & \cmark& ~ &~ & ~ & ~&~\\
         % \midrule
         \citep{holden2016deep,holden2017fast}&~ & ~& ~ &\cmark & ~&~&~ \\
         % \midrule
         \citep{aberman2020unpaired}& \cmark & ~& ~ &~ & ~&~&~\\
         % \midrule
         % \citep{wen2021autoregressive}
         \citep{park2021diverse}& \cmark & \cmark& ~ &~ & ~&~&\cmark\\
         % \midrule
        \citep{tao_style-erd_2022}& \cmark & ~& ~ &~ & ~&~&~\\

         \citep{jang2022motion}& ~ & ~& ~ &\cmark & ~&~&~\\
         % \midrule
         \midrule
         Ours&\cmark & \cmark& ~ &\cmark & \cmark&~&\cmark \\
         \bottomrule
    \end{tabular}
    }
    \caption{\footnotesize{Our generative framework owns flexible design for training and inference.}}
    \label{tab:flexibility}
    \vspace{-1.8em}
\end{table}
    % \vspace{1.0em}

\paragraph{Motion Style Transfer.}
Motion style transfer has been a long-standing challenge in computer animation. 
% In order to manipulate styles, early machine learning approaches heavily depend on delicate-designed features in the temporal domain~\citep{amaya1996emotion} or the spectrum~\citep{unuma1995fourier}. 
%The rising of data-driven methods that spontaneously learn from examples gets the elusive motion style free from precise mathematical definitions. 
% The linear time-invariant (LTI) model is proposed based on per-frame pairwise between motions with similar content but heterogeneous styles~\citep{hsu2005style}. To alleviate the demand for densely paired content, 
%For example, the mixtures of autoregressive (MAR) models following a KNN search are designed to organize motion data more efficiently for online style transfer~\citep{xia2015realtime}. 
Early work~\citep{xia2015realtime} design an online style transfer system based on KNN search. ~\citep{holden2016deep,du2019stylistic,yumer_spectral_2016} transfers the style from reference to source motion through optimizing style statistic features, such as Gram matrix, which are computationally intensive. Feed-forward based approaches~\citep{holden2017fast,aberman2020unpaired,smith_efficient_2019} properly address this problem, where ~\citep{aberman2020unpaired} finalizes a two-branch pipeline based on deterministic autoencoders and AdaIN~\citep{huang2017arbitrary} for style-content disentanglement and composition; while ~\citep{smith_efficient_2019} manages to stylize existing motions using one-hot style label, and models it as an class conditioned generation process. More recently, with the explosion of deep learning techniques, some works adopt graph neural networks (GNN)~\citep{park2021diverse, jang2022motion}, advanced time-series model ~\citep{tao_style-erd_2022, wen2021autoregressive}, or diffusion model~\citep{raab2023single} to the motion style transfer task. Specifically, ~\citep{jang2022motion} realizes a framework that extracts style features from motion body parts. %Meanwhile, the method from ~\citep{raab2023single} uses diffusion model to learn the internal motifs from a single motion sequence. 
%These existing works primarily learn an unconstrained style space for deterministic one-to-one style mapping. This is in the contrast with our generative style space that is capable of generating diverse and novel stylization results.

\paragraph{Synthesis in Latent.} Deep latent have been evidenced as a promising alternative representation for content synthesis including images~\citep{rombach2022high,ramesh2022hierarchical,esser2021taming,esser2021imagebart,ding2021cogview}, motion~\citep{guo2022generating,guo2022tm2t,gong2023tm2d,chen2022executing}, 3D shape~\citep{zeng2022lion,fu2022shapecrafter}, and video~\citep{yan2021videogpt,hu2023lamd}. These works commonly adopt a two-stage synthesis strategy. At the first stage, the source contents are encoded into continuous latent maps (\textit{e.g.}, using autoencoders, CLIP~\citep{radford2021learning}), or discrete latent tokens through VQ-VAE~\citep{van2017neural}. Then, models are learned to generate these latent representation explicitly which can be inverted to data space in the end. This strategy has shown several merits. Deep latent consists of the most representative features of raw data, which leads to a more expressive and compact representation. It also cuts down the cost of time and computation during training and inference. These prior arts inspire the proposed latent stylization in our approach.

\section{Generative Motion Stylization}

%Given a content motion and style cues such as a style motion or a label $sl\in\{1,...,N\}$, where $N$ denotes the number of styles, our goal is to synthesize a motion sequence which exhibits the same \textit{action semantics} as presented in the content motion, as well as conveying the \textit{style} provided by style clues. 
%Content motions have fixed length during training, while at test time it can be in arbitrary length. 
An overview of our method is described in~\Cref{fig:model}. Motions are first projected into the latent space (\cref{subsec:motion_latent}). With this, the latent stylization framework learns to extract the content and style from the input code (\cref{subsec:stylization}), which further support multiple applications during inference (\cref{subsec:applications}).

\vspace{-0.5em}
\subsection{Motion Latent Representation}
\label{subsec:motion_latent}

As a pre-processing step, we learn a motion autoencoder that builds the mapping between motion and latent space. More precisely, given a pose sequence $\mathbf{P}\in\mathbb{R}^{T\times D}$, where $T$ denotes the number of poses and $D$ pose dimension, the encoder $\mathcal{E}$ encodes $\mathbf{P}$ into a motion code $\mathbf{z}=\mathcal{E}(\mathbf{P}) \in \mathbb{R}^{T_z\times D_z}$, with $T_z$ and $D_z$ the temporal length and spatial dimension respectively, and then the decoder $\mathcal{D}$ recovers the input motion from the latent features, formally $\hat{\mathbf{P}} = \mathcal{D}(\mathbf{z}) = \mathcal{D}(\mathcal{E}(\mathbf{P}))$. 

A well-learned latent space should exhibit smoothness and low variance. In this work, we experiment with two kinds of regularization methods in latent space: 1) as in VAE~\citep{kingma2013auto}, the latent space is formed under a light KL regularization towards standard normal distribution $\mathcal{L}^l_{kld} = \lambda^l_{kld}D_\mathrm{KL}(\mathbf{z}||\mathcal{N}(\mathbf{0},\mathbf{I}))$ ; and 2) similar to~\citep{guo2022generating}, we train the classical autoencoder and impose L1 penalty on the magnitude and smoothness of motion code sequences, giving $\mathcal{L}^l_{reg} = \lambda_{l1}\|\mathbf{z}\|_1 + \lambda_{sms}\|\mathbf{z}_{1:T_z}-\mathbf{z}_{0:T_z-1}\|_1$. Our motion encoder $\mathcal{E}$ and decoder $\mathcal{D}$ are simply 1-D convolution layers with downsampling and upsampling scale of 4 (\textit{i.e.,} $T=4T_z$), resulting in a more compact form of data that captures temporal semantic information. 
%Moreover, this autoencoding can be learned on the large unlabeled human motion dataset, and then reused for various annotated datasets which are usually in smaller scale.

\subsection{Motion Latent Stylization Framework}
\label{subsec:stylization}
\begin{figure*}[t]
	\centering
	\includegraphics[width=0.95\linewidth]{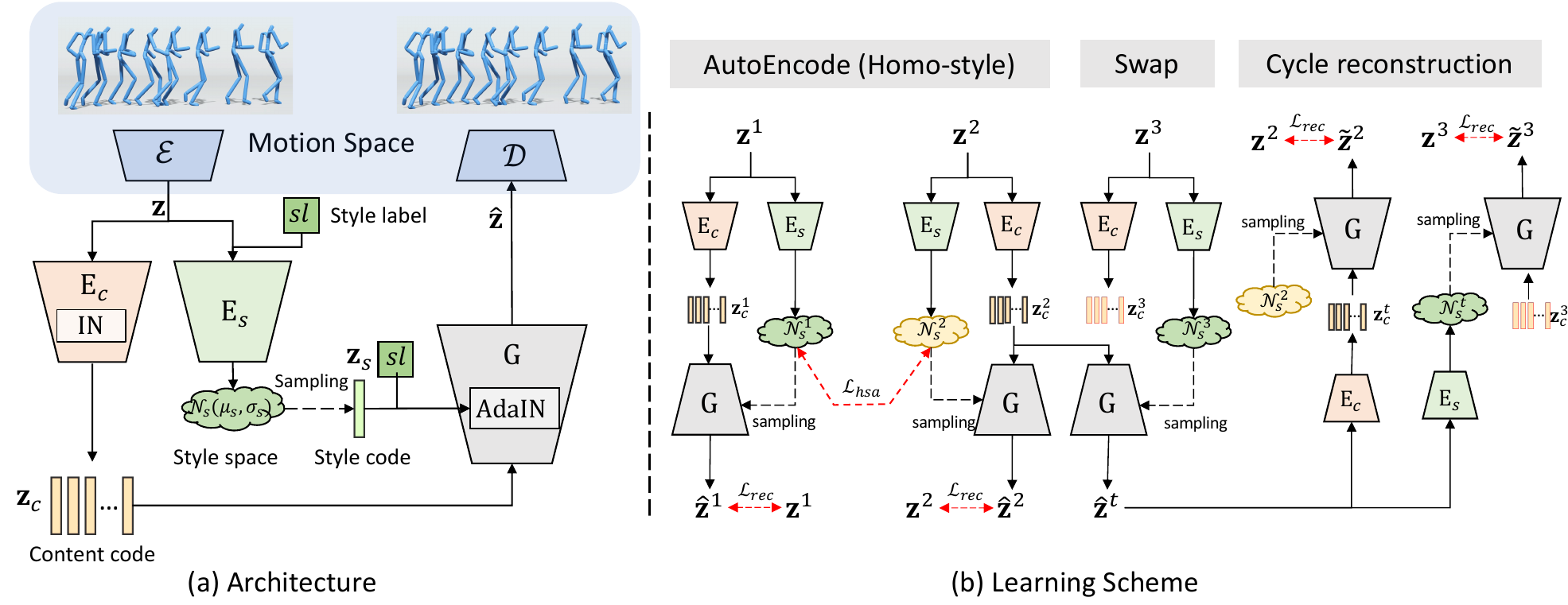}
 	 \vspace{-1em}
	\caption{\footnotesize{\textbf{Approach overview.} (a) A pre-trained autoencoder $\mathcal{E}$ and $\mathcal{D}$ (\cref{subsec:motion_latent}) builds the mappings between \textit{motion} and \textit{latent} spaces. Motion (latent) code $\mathbf{z}$ is further encoded into two parts: content code $\mathbf{z}_c$ from content encoder ($\mathrm{E}_c$), and style space $\mathcal{N}_s$ from style encoder ($\mathrm{E}_s$) that take style label $sl$ as an additional input. The content code ($\mathbf{z}_c$) is decoded back to motion code ($\mathbf{\hat{z}}$) via generator $\mathrm{G}$. Meanwhile, a style code $\mathbf{z}_s$ is sampled from style space ($\mathcal{N}_s$), together with style label ($sl$), which are subsequently injected to generator layers through adaptive instance normalization (AdaIN). (b) Learning scheme, where style label ($sl$) is omitted for simplicity. Our model is trained by autoencoding for content and style coming from the \textbf{same} input. When decoding with content from \textbf{different} input (\textit{i.e.,} swap), we enforce the resulting motion code ($\mathbf{\hat{z}}^t$) to follow the cycle reconstruction constraint. For motion codes ($\mathbf{z}^1$, $\mathbf{z}^2$) segmented from the same sequence (homo-style), their style spaces are assumed to be close and learned with style alignment loss $\mathcal{L}_{hsa}$.}}
	\label{fig:model}
	 \vspace{-1.5em}
\end{figure*}

As depicted in~\Cref{fig:model}, our latent stylization framework aims to yield a valid parametric style space, and meanwhile, preserve semantic information in content codes as much as possible. This is achieved by our specific model design and dedicated learning strategies. 

\vspace{-1em}
\subsubsection{Model Architecture.}

There are three principal components in our framework: a style encoder $\mathrm{E}_s$, a content encoder $\mathrm{E}_c$ and a generator $\mathrm{G}$, as in \Cref{fig:model} (a). 

\revise{\noindent\textbf{Probabilistic Style Space.}} For style, existing works~\citep{park2021diverse,aberman2020unpaired,jang2022motion} generate deterministic style code from motion input. In contrast, our style encoder $\mathrm{E}_s$, taking $\mathbf{z}$ and style label $sl$ as input, produces a vector Gaussian distribution $\mathcal{N}_s(\mu_s,\sigma_s)$ to formulate the style space, from which a style code $\mathbf{z}_s\in\mathbb{R}^{D_z^s}$ is sampled. \revise{In test-time, this probabilistic style space enables us to generate diverse and novel style samples.}

\revise{Comparing to style features, content features exhibit more locality and determinism. Therefore, we model them deterministically to preserve the precise structure and meaning of the motion sequence.} The content encoder converts the a motion code $\mathbf{z}\in\mathbb{R}^{T_z\times D_z}$ into a content code $\mathbf{z}_c\in\mathbb{R}^{T_z^c\times D_z^c}$ that keeps a temporal dimension $T_z^c$, where global statistic features (style) are erased through instance normalization (IN). The asymmetric shape of content code $\mathbf{z}_c$ and style code $\mathbf{z}_s$ are designed of purpose. We expect the former to capture local semantics while the latter encodes global features, as what style is commonly thought of. Content code is subsequently fed into the convolution-based generator $\mathrm{G}$, where the mean and variance of each layer output are modified by an affine transformation of style information (\textit{i.e.,} style code and label), known as adaptive instance normalization (AdaIN). The generator aims to transform valid combinations of content and style into meaningful motion codes in the latent space.

\vspace{-0.5em}
\subsubsection{Learning Scheme} 
\label{subsubsec:learning_scheme}

With the model mentioned above, we propose a series of strategies for learning disentangled content and style representations. ~\Cref{fig:model} (b) illustrates our learning scheme. Note the input of style label $sl$ is omitted for simplicity. During training, for each iteration, we design three groups of inputs: $\mathbf{z}^1$, $\mathbf{z}^2$ and $\mathbf{z}^3$, where $\mathbf{z}^1$ and $\mathbf{z}^2$ are motion code segments coming from the same sequence and $\mathbf{z}^3$ can be any other segments. 

\begin{figure*}[t]
	\centering
	\includegraphics[width=0.8\linewidth]{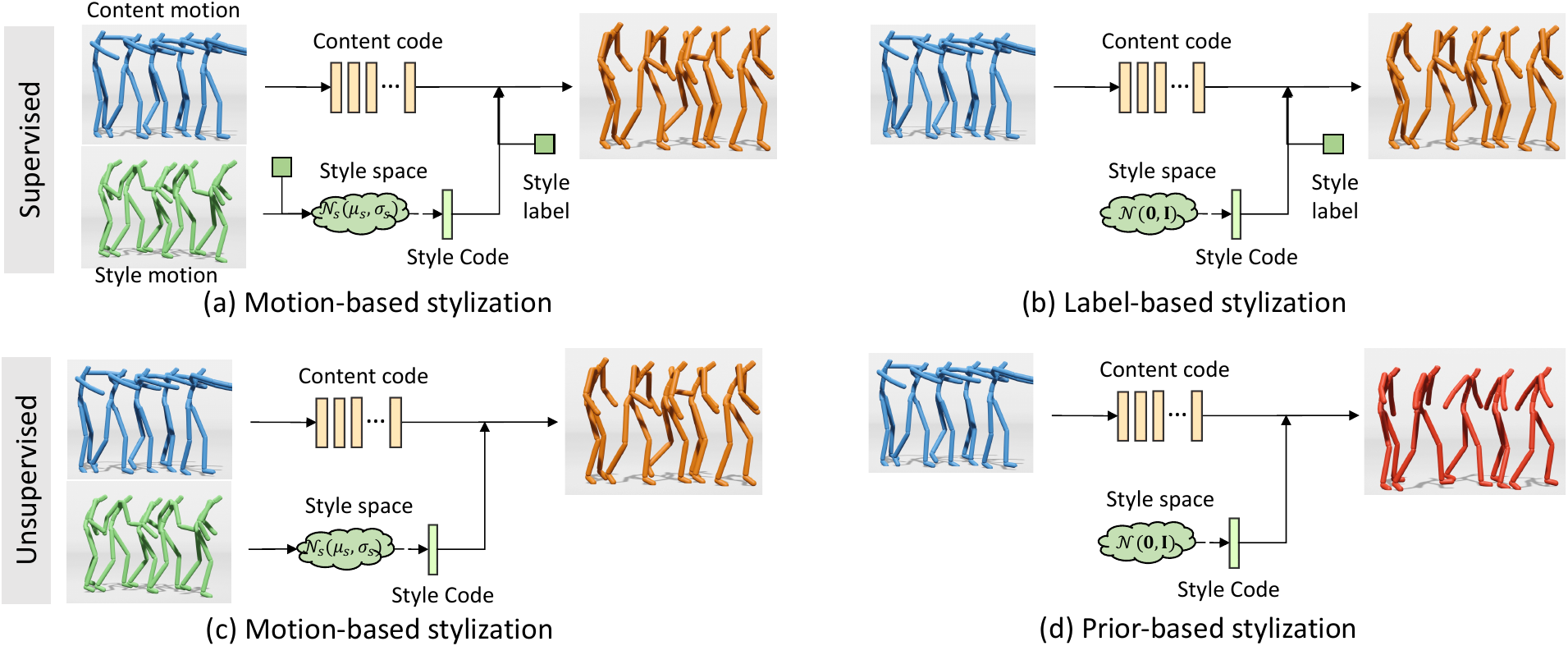}
 	 \vspace{-1em}
	\caption{\footnotesize{During inference, our approach can stylize input content motions with the style cues from (a, c) motion, (b) style label and (d) unconditional style prior space.}}
	\label{fig:inference}
	 \vspace{-1.5em}
\end{figure*}

\paragraph{AutoEncoding $\mathcal{L}_{rec}$.} We train our latent stylization framework partly through autoencoding, that given motion latent codes, like $\mathbf{z}^1$ and $\mathbf{z}^2$, the generator learns to reconstruct the input from the corresponding encoded content and style features, formally $\mathbf{\hat{z}} = \mathrm{G}(\mathrm{E}_c(\mathbf{z}),\mathrm{E}_s(\mathbf{z}))$. For accurate reconstruction, we decode the resulting motion latent codes ($\mathbf{\hat{z}}^1$ and $\mathbf{\hat{z}}^2$) back to motion space ($\mathbf{\hat{P}}^1$ and $\mathbf{\hat{P}}^2$) through $\mathcal{D}$, and apply L1-distance reconstruction in both latent and motion space:
\begin{small}
% \small
\begin{align}
    \mathcal{L}_{rec} = \small{\sum_{i\in \{1,2\}}}\|\mathbf{\hat{z}}^i - \mathbf{z}^i\|_1 + \|\mathbf{\hat{P}}^i - \mathbf{P}^i\|_1
\end{align}
\vspace{-2em}
\end{small}

\paragraph{Homo-style Alignment $\mathcal{L}_{hsa}$.} For the motion segments in one motion sequence, we could usually assume their styles are similar in all aspects. This is a strong supervision signal especially when style annotation is unavailable, dubbed \textit{homo-style alignment} in our work. Since $\mathbf{z}^1$ and $\mathbf{z}^2$ belong to the same sequence, their learned style spaces are enforced to be close:
\begin{small}
\begin{align}
    \mathcal{L}_{hsa} = D_\mathrm{KL}(\mathcal{N}_s^1(\mu_s^1,\sigma_s^1)\|\mathcal{N}_s^2(\mu_s^2,\sigma_s^2))
\end{align}
\end{small}
\vspace{-2.5em}

% \yxmu{\[ \mathcal{L}_{hsa} = \mathbb{E}_{\mathbf{z}^{1}, \mathbf{z}^{2} \sim p(\cdot)}\left [\mathrm{KL} \left (q_{E_s} (\mathbf{z}^{1}_{s}|\cdot)||q_{E_s}(\mathbf{z}^{2}_{s}|\cdot) \right )\right ], \] where we ignore the mapping $q_\mathcal{E}(\mathbf{z}^{1,2}|\mathbf{P}^{1,2})$ for simplicity.}

\paragraph{Swap and Cycle Reconstruction $\mathcal{L}_{cyc}$.} To further encourage content-style disentanglement, we adopt a cycle consistency constraint~\citep{zhu2017unpaired,jang2022motion} when content and style are swapped between different motion codes, such as $\mathbf{z}^2$ and $\mathbf{z}^3$ in \cref{fig:model}. Specifically, the generator $\mathrm{G}$ takes as input the content from $\mathbf{z}^2$ and the style from $\mathbf{z}^3$, and then produces a new \textit{transferred} motion code $\mathbf{z}^t$, which are supposed to preserve the content information from $\mathbf{z}^2$ and the style from $\mathbf{z}^3$. Therefore, if we re-combine $\mathbf{z}^t$'s content and $\mathbf{z}^2$'s style, the generator should be able to restore $\mathbf{z}^2$. The same to $\mathbf{\tilde{z}}^3$ that are recovered from the mix of $\mathbf{z}^t$'s style and $\mathbf{z}^3$'s content :
\begin{small}
    \begin{align}
    \mathcal{L}_{cyc} = \sum_{i\in \{2, 3\}}\|\mathbf{\tilde{z}}^i - \mathbf{z}^i\|_1  + \|\mathbf{\tilde{P}}^i - \mathbf{P}^i\|_1 
\end{align}
\end{small}

\vspace{-1.5em}

% \yxmu{\[ \mathcal{L}_{cyc} = \mathbb{E}_{\mathbf{z}_{s}, \mathbf{z}_{c} \sim q_{E_{s}, E_{c}}(\cdot | \mathbf{z}^{2}, \mathbf{z}^{t})}[-\ln p_{G}(\mathbf{z}^{2}|\cdot) - \ln p_{G, \mathcal{D}}(\mathbf{P}^{2}|\cdot)]
% + \mathbb{E}_{\mathbf{z}_{s}, \mathbf{z}_{c} \sim q_{E_{s}, E_{c}}(\cdot | \mathbf{z}^{t}, \mathbf{z}^{3})}[-\ln p_{G}(\mathbf{z}^{3}|\cdot) - \ln p_{G, \mathcal{D}}(\mathbf{P}^{3}|\cdot)], \] where $\mathbf{z}^{t} \sim p_{G}(\cdot|\mathbf{z}_{s}^{3}, \mathbf{z}_{c}^{2})$. 
% or omit the expectation estimation, just take the L1 loss when calculate loss for likelihood, e.g. formula (1)(3)
% }

To ensure smooth and samplable style spaces, we apply a KL loss regularization to all style spaces:
% \small{
\begin{small}
\begin{align}
    \mathcal{L}_{kl} = \sum_{i\in\{1, 2, 3, t\}}D_\mathrm{KL}(\mathcal{N}_s^i(\mu_s^i,\sigma_s^i))\|\mathcal{N}(\mathbf{0}, \mathbf{I}))
\end{align}    
\end{small}

\vspace{-1em}
% \yxmu{Maybe just take: $\[ \mathcal{L}_{kl} =  \sum_{i}^{\{1,2,3,t\}}\mathbb{E}_{\mathbf{z}^{i} \sim p(\cdot)}[\mathrm{KL}(q_{E_s} (\mathbf{z}^{i}_{s}|\cdot)\|\mathcal{N}(\mathbf{0}, \mathbf{I}))], \]$}

Overall, our final objective is $\mathcal{L} = \mathcal{L}_{rec} + \lambda_{hsa}\mathcal{L}_{hsa} + \lambda_{cyc}\mathcal{L}_{cyc}+\lambda_{kl}\mathcal{L}_{kl}$. We also have experimented adversarial loss for autoencoding and cycle reconstruction as in~\citep{park2021diverse,aberman2020unpaired,tao_style-erd_2022}, which however appears to be extremely unstable in training.

\paragraph{Unsupervised Scheme (\textit{w/o} Style Label).} Collecting style labeled motions is resource consuming. Our approach can simply fit in unsupervised setting with just one-line change of code during training---to drop out style label $sl$ input. %All assumptions made above, \textit{e.g.}, homo-style alignment, are still applicable.

\paragraph{Difference of $\mathcal{N}_s$ Learned \textit{w} and \textit{w/o} Style Label.} While learning with style label, since both the style encoder $\mathrm{E}_s$ and generator $\mathrm{G}$ are conditioned on style label, the style space is encouraged to learn style variables other than style label as illustrated in~\cref{fig:latent_vis} (d). Whereas in unsupervised setting where the networks are agnostic to style label, in order to precisely reconstruct motions, the style space is expected to cover the \textit{holistic} style information, including style label (see~\cref{fig:latent_vis} (c)).

\subsubsection{Global Motion Prediction}
\label{subsec:gmp}
% The root velocity, also known as global motion, is usually a more nuanced factor than local joint motions. Small deviations in global motion can greatly compromise the motion quality, such as foot sliding. However, the precise correspondence of global motions between content and style motions are hard to learn without paired data. 
Global motion (\textit{i.e.,} root velocity) is perceptually a more sensitive element than local joint motion (e.g., foot skating).
%, where minor deviations can significantly harm motion naturalness, causing issues like foot sliding. 
% However, learning transferred global motion from the input content and style motions is challenging without paired data. \yxmu{TODO}
However, given one motion, transferring its global motion to another style domain is challenging without supervision of paired data.
Previous works commonly calculate the target global motion directly from the content motion, or enforce them to be close in training. This may fail when the transferred motion differs a lot from the source content. In our work, we propose a simple yet effective alternative, which is a small 1D convolutional network that predicts the global motion from local joint motion, simply trained on unlabeled data using objective of mean absolute error. 
%To enhance robustness, the input local motion is added with random noise in training. 
During inference, the global motion of output can be accurately inferred from its local motion.

\vspace{-0.5em}
\subsection{Inference Phase}
\label{subsec:applications}
As displayed in \Cref{fig:inference}, our approach at run time can be used in multiple ways. In \textit{supervised} setting: a) \textbf{motion-based} stylization requires the user to provide a style motion and a style label as the style references; and b) \textbf{label-based} stylization only asks for a target style label for stylization. With sampled style codes from a standard normal distribution $\mathcal{N}(\mathbf{0}, \mathbf{I})$, we are able to stylize source content motion non-deterministically. In the case of \textit{unsupervised} setting: c) motion-based stylization, which similarly, yields a style code from a reference motion; and d) \textbf{prior-based} stylization that samples random style codes from the prior distribution $\mathcal{N}(\mathbf{0}, \mathbf{I})$. Since there is no other pretext style indications, the output motion could carry any style trait in the style space.

\section{Experiments}
\label{sec:experiment}

%We carry out extensive experiments on various datasets using a set of evaluation metrics.
% \subsection{Datasets}
We adopt three datasets for comprehensive evaluation. ~\citep{aberman2020unpaired} is a widely used motion style dataset, which contains 16 distinct style labels including \textit{angry}, \textit{happy}, \textit{Old}, etc, with total duration of 193 minute. ~\citep{xia2015realtime} is much smaller motion style collection (25 mins) that is captured in 8 styles, with accurate action type annotation (8 actions). The motions are typically shorter than 3s. The other one is CMU Mocap~\citep{cmu2021mocap}, an unlabeled dataset with high diversity and quantity of motion data. All motion data is retargeted to the same 21-joint skeleton structure, with a 10\% held-out subset for evaluation. Our autoencoders and global motion regressor are trained on the union of all training sets, while the latent stylization models are trained \textbf{excursively} on~\citep{aberman2020unpaired}, using the other two for zero-shot evaluation. During evaluation, we use the styles from~\citep{aberman2020unpaired} test sets to stylize the motions from one of the three test sets. Style space is learned based on motions of 160 poses (5.3s). Note our models supports stylization of arbitrary-length content motions. See~\Cref{sec:implementation} for implementation details and full model architectures.
\paragraph{Metrics} in previous motion stylization works heavily rely on a sparse set of measurements, typically human evaluation and style accuracy. Here, we design a suite of metrics to comprehensively evaluate our approach. We firstly pre-train a style classifier on~\citep{aberman2020unpaired} train set, and use it as a style feature extractor to compute \textit{style recognition accuracy} and \textit{style FID}. For dataset with available action annotation (~\citep{xia2015realtime}), an action classifier is learned to extract content features and calculate \textit{content recognition accuracy} and \textit{content FID}. We further evaluate the content preservation using \textit{geodesic distance} of the local joint rotations between input content motion and generated motion. \textit{Diversity} in ~\citep{lee2019dancing} is also employed to quantify the stochasticity in the stylization results. Further explanations are provided in~\cref{sec:metric}. 

\paragraph{Baselines.} We compare our method to three state-of-the-art methods~\citep{aberman2020unpaired,jang2022motion,park2021diverse} in their respective settings. Among these, \citep{aberman2020unpaired} and \citep{park2021diverse} are supervised methods learned within GAN framework. %~\citep{aberman2020unpaired} utilizes style label to constrain style space in training, and takes reference motion as style input during inference. 
\citep{park2021diverse} learns per-label style space, and a mapping between Gaussian space and style space. At run time, it supports both deterministic motion-based and diverse label-based motion stylization. %Note recent work ~\citep{tao_style-erd_2022} is not included as their approach requires action labels for supervision, which are unavailable in our context . See~\Cref{subsec:baseline} for baseline implementation details.

%It takes around 12 hours to train our model on a single Tesla P100 16G GPU.

% \subsection{Results}
% \label{subsec:comparison}

\begin{table}[t]
    \centering
    \scalebox{0.55}{
    \begin{tabular}{l c c c c c c c c c c}
    \toprule
    % \multirow{2}{*}{Methods} & \multicolumn{4}{c}{\DN (Coarse-grained)} & & \multicolumn{4}{c}{HumanAct(Fine-grained)} \\
    % \cline{2-5}
    % \cline{7-10}
    %                 & FID$\downarrow$ & Accuracy$\uparrow$ & Diversity$\rightarrow$& MModality$\rightarrow$ &  & FID$\downarrow$ & Accuracy$\uparrow$ & Diversity$\rightarrow$ & MModality$\rightarrow$\\
    \multirow{2}{*}{Setting}  & \multirow{2}{*}{Methods}& \multicolumn{4}{c}{~\citep{aberman2020unpaired}} & ~ &\multicolumn{4}{c}{CMU Mocap~\citep{cmu2021mocap}} \\
    \cline{3-6}
    \cline{8-11}
    ~ & ~& Style Accuracy$\uparrow$ & Style FID$\downarrow$ & Geo Dis$\downarrow$ & Div$\uparrow$& ~& Style Accuracy$\uparrow$ & Style FID$\downarrow$ & Geo Dis$\downarrow$ & Div$\uparrow$ \\
    \midrule
    ~& Real Motions & \et{0.997}{002} & \et{0.002}{000} & - & - & ~&\et{0.997}{002} & \et{0.002}{000} & - & -\\
    \midrule
    \multirow{5}{*}{Motion-based (S)} & ~\citep{aberman2020unpaired} & \et{0.547}{016} & \et{0.379}{018} & \et{0.804}{003} & - & ~&\et{0.445}{009} & \et{0.508}{011} & \et{0.910}{002} & -\\
    ~ & ~\citep{park2021diverse} & \et{0.891}{007} & \et{0.038}{003} & \et{0.531}{001} & - & ~&\et{0.674}{014} & \et{0.136}{011} & \et{0.663}{003} & -\\
    & Ours w/o latent & \et{0.932}{008} & \et{0.022}{002} & \et{0.463}{003} & - & ~&\et{0.879}{008} & \et{0.046}{004} & \et{0.636}{004} & -\\
    & Ours (V) & \ets{0.935}{007} & \ets{0.020}{002} & \ets{0.426}{003} & - & ~&\ets{0.918}{010} & \etb{0.028}{003} & \ets{0.629}{002} & -\\
    & Ours (A) & \etb{0.945}{007} & \etb{0.020}{002} & \etb{0.344}{002} & - & ~&\etb{0.918}{007} & \ets{0.031}{003} & \etb{0.569}{002} & -\\
    
    \midrule
\multirow{4}{*}{Label-based (S)} & ~\citep{park2021diverse} & \etb{0.971}{006} & \etb{0.013}{001} & \et{0.571}{002} & \et{0.146}{009} & ~&\et{0.813}{010} & \et{0.065}{007} & \et{0.693}{004} & \et{0.229}{019}\\
    ~& Ours w/o latent  & \et{0.933}{009} & \et{0.023}{002} & \et{0.447}{002} & \etb{0.174}{017} & ~&\et{0.882}{008} & \et{0.053}{003} & \ets{0.611}{003} & \etb{0.266}{021}\\
    ~& Ours (V)  & \ets{0.946}{007} & \et{0.020}{002} & \ets{0.427}{003} & \ets{0.134}{016} & ~&\etb{0.923}{007} & \etb{0.027}{003} & \et{0.614}{002} & \ets{0.193}{013}\\
    ~& Ours (A)  & \et{0.942}{006} & \ets{0.019}{001} & \etb{0.344}{003} & \et{0.050}{006} & ~&\ets{0.915}{005} & \ets{0.031}{003} & \etb{0.571}{003} & \et{0.067}{005}\\
    \midrule
    
    \multirow{4}{*}{Motion-based (U)} & ~\citep{jang2022motion} & \ets{0.833}{010} & \et{0.047}{004} & \et{0.559}{003} & - & ~&\et{0.793}{009} & \et{0.058}{004} & \et{0.725}{004} & -\\
    & Ours w/o latent & \et{0.780}{014} & \et{0.048}{003} & \ets{0.466}{004} & - & ~&\et{0.761}{009} & \et{0.082}{005} & \etb{0.645}{003} & -\\
    & Ours (V) & \etb{0.840}{010} & \etb{0.036}{003} & \et{0.478}{004} & - & ~&\etb{0.828}{010} & \etb{0.052}{004} & \et{0.672}{003} & -\\
    & Ours (A) & \et{0.804}{011} & \ets{0.040}{003} & \etb{0.441}{003} & - & ~&\ets{0.799}{009} & \ets{0.056}{003} & \ets{0.648}{004} & -\\
    \midrule
    \multirow{3}{*}{Prior-based (U)} & Ours w/o latent & - & - & \ets{0.431}{003} & \ets{1.169}{030} & ~& - & - & \ets{0.626}{001} & \ets{1.252}{029}\\
    ~& Ours (V) & - & - & \etb{0.418}{003} & \et{1.069}{028} & ~& - & - & \et{0.611}{003} & \et{0.857}{024}\\
    ~& Ours (A)& - & - & \et{0.436}{004} & \etb{1.187}{029} & ~& - & - & \etb{0.641}{002} & \ets{0.949}{022}\\ 
    \bottomrule
    \end{tabular}
    }
    \caption{\footnotesize{Quantitative results on the ~\citep{aberman2020unpaired} and CMU Mocap test sets. $\pm$ indicates 95\% confidence interval. \textbf{Bold} face indicates the best result, while \underline{underscore} refers to the second best. (S) and (U) denote \textit{supervised} and \textit{unsupervised} setting. (V) VAE and (A) AE represent different latent models in~\cref{subsec:motion_latent}}.}
    \label{tab:aberman_cmu}
    	 % \vspace{-1em}

\end{table}

\begin{table}[t]
% \centering
\begin{minipage}[b]{0.6\linewidth}
% \centering
    \scalebox{0.53}{
    \begin{tabular}{l c c c c c c}
    \toprule
    % \multirow{2}{*}{Methods} & \multicolumn{4}{c}{\DN (Coarse-grained)} & & \multicolumn{4}{c}{HumanAct(Fine-grained)} \\
    % \cline{2-5}
    % \cline{7-10}
    %                 & FID$\downarrow$ & Accuracy$\uparrow$ & Diversity$\rightarrow$& MModality$\rightarrow$ &  & FID$\downarrow$ & Accuracy$\uparrow$ & Diversity$\rightarrow$ & MModality$\rightarrow$\\
    \multirow{2}{*}{Setting}  & \multirow{2}{*}{Methods}& \multicolumn{4}{c}{~\citep{xia2015realtime}} & \\
    \cline{3-7}
    ~ & ~& Style Acc$\uparrow$ & Content Acc$\uparrow$  & Content FID$\downarrow$ & Geo Dis$\downarrow$ & Div$\uparrow$\\
    \midrule
    % ~& Real Motions & - & \et{0.953}{003} & - & - & - \\
    % \midrule
    \multirow{5}{*}{M-based (S)} &~\citep{aberman2020unpaired} & \et{0.364}{011} & \et{0.318}{008} & \et{0.705}{014} & \et{0.931}{003} & - \\
    &  ~\citep{park2021diverse}& \et{0.527}{006} & \et{0.441}{009} & \et{0.381}{010} & \ets{0.698}{001} & - \\
    &Ours w/o latent & \et{0.851}{012} & \ets{0.654}{012} & \et{0.258}{007} & \et{0.707}{004} & - \\
    & Ours (V)& \etb{0.934}{006} & \et{0.579}{016} & \ets{0.210}{004} & \et{0.716}{003} & - \\
    & Ours (A)& \ets{0.926}{008} & \etb{0.674}{011} & \etb{0.189}{005} & \etb{0.680}{003} & - \\
    \midrule
    \multirow{4}{*}{L-based (S)} & ~\citep{park2021diverse} & \et{0.796}{007} & \et{0.311}{009} & \et{0.507}{011} & \et{0.770}{003} & \et{0.175}{014} \\
    &Ours w/o latent & \et{0.843}{012} & \ets{0.655}{013} & \et{0.264}{008} & \ets{0.691}{003} & \etb{0.281}{032} \\
    & Ours (V)& \etb{0.944}{008} & \et{0.606}{013} & \ets{0.208}{005} & \et{0.705}{003} & \ets{0.228}{023} \\
    & Ours (A)& \ets{0.933}{011} & \etb{0.668}{014} & \etb{0.193}{005} & \etb{0.679}{002} & \et{0.095}{013} \\
    \midrule
    \multirow{4}{*}{M-based (U)} & ~\citep{jang2022motion} & \et{0.658}{009} & \et{0.337}{017} & \et{0.380}{011} & \et{0.857}{004} & - \\
    &Ours w/o latent & \et{0.734}{014} & \ets{0.584}{011} & \et{0.272}{008} & \etb{0.721}{003} & - \\
    & Ours (V)& \etb{0.860}{010} & \et{0.499}{015} & \ets{0.221}{006} & \et{0.747}{004} & - \\
    & Ours (A)& \ets{0.814}{011} & \etb{0.588}{010} & \etb{0.217}{006} & \ets{0.735}{003} & - \\
    \midrule
    \multirow{3}{*}{P-based (U)} & Ours w/o latent & - & \etb{0.627}{014} & \et{0.246}{007} & \ets{0.708}{003} & \etb{1.193}{029} \\
    & Ours (V) & - & \et{0.579}{013} & \ets{0.239}{006} & \etb{0.704}{002} & \et{0.874}{029} \\
    & Ours (A) & - & \ets{0.586}{015} & \etb{0.227}{006} & \et{0.736}{003} & \ets{0.978}{026} \\
    \bottomrule
    \end{tabular}
    }
    \caption{\footnotesize{Quantitative results on the ~\citep{xia2015realtime} test set.}}
    \label{tab:xia}
\end{minipage}
% \hspace{0.7cm}
\hspace{0.063\linewidth}
\begin{minipage}[b]{0.32\linewidth}
\centering
    \scalebox{0.6}{
    \begin{tabular}{l c c}
    \toprule

    Setting & $\text{Method}$ & $\text{Ours wins}$\\
    \midrule
    % ~& Real Motions & - & \et{0.953}{003} & - & - & - \\
    % \midrule
    \multirow{3}{*}{M-based (S)} & ~\citep{aberman2020unpaired} ~ & 78.69\% \\
    % ~ & 21.31\% & 78.69\% \\
    % \cline{2-3}
    ~ & ~\citep{park2021diverse} & 73.67\% \\
    % ~ & 26.33\% & 73.67\% \\
    % \cline{2-3}
    ~ & Ours w/o latent & 65.98\% \\
    % ~ & 34.02\% & 65.98\% \\
    \midrule
    L-based (S) & ~\citep{park2021diverse} & 73.06\% \\
    % ~ & 26.94\% & 73.06\% \\
    \midrule
    M-based (U) & ~\citep{jang2022motion} & 58.92\% \\
    % ~ & 41.08\% & 58.92\% \\
    \bottomrule
    \end{tabular}
    }
    \caption{\footnotesize{Human evaluation results.}}
    \label{tab:userstudy}

    \vspace{0.15cm}

    \scalebox{0.6}{
    \begin{tabular}{c c c}
    \toprule
    % \multirow{2}{*}{Methods} & \multicolumn{4}{c}{\DN (Coarse-grained)} & & \multicolumn{4}{c}{HumanAct(Fine-grained)} \\
    % \cline{2-5}
    % \cline{7-10}
    %                 & FID$\downarrow$ & Accuracy$\uparrow$ & Diversity$\rightarrow$& MModality$\rightarrow$ &  & FID$\downarrow$ & Accuracy$\uparrow$ & Diversity$\rightarrow$ & MModality$\rightarrow$\\
    Methods & Runtime (ms)$\downarrow$\\
    \midrule
    % ~& Real Motions & - & \et{0.953}{003} & - & - & - \\
    % \midrule
    ~\citep{aberman2020unpaired} & 16.763 \\
    ~\citep{park2021diverse}(M) & 37.247\\
    ~\citep{park2021diverse}(L) & \underline{16.329}\\
    ~\citep{jang2022motion} & 67.563\\
    % Ours w/o latent & \textbf{4.682} & 13.51M\\
    % Ours (V)& 5.026 & 19.21M\\
    Ours (A)(M)& \textbf{4.760}\\
    \bottomrule
    \end{tabular}
    }
    % \begin{minipage}{0.8\textwidth}
        \caption{\footnotesize{Runtime comparisons.}}
    % \end{minipage}
    
    \label{tab:infertime}
\end{minipage}
	 \vspace{-2em}
\end{table}

\subsection{Quantitative Results}
\label{subsec:quantitative_results}
\Cref{tab:aberman_cmu} and \Cref{tab:xia} present the quantitative evaluation results on the test sets of ~\citep{aberman2020unpaired}, CMU Mocap~\citep{cmu2021mocap} and ~\citep{xia2015realtime}. Note the latter two datasets are completely unseen to our latent stylization models. We generate results using motions in these three test sets as content, and randomly sample style motions and labels from ~\citep{aberman2020unpaired} test set. For fair comparison, we repeat this experiment 30 times, and report the mean value with a 95\% confidence interval. We also consider the variants of our approach: non-latent stylization (\textit{ours w/o latent}), using VAE (\textit{Ours (V)}) or AE (\textit{ Ours (A)}) as latent model (See \cref{subsec:motion_latent}). \revise{\textit{Ours w/o latent} employs the identical architecture as our full model, as illustrated in~\cref{fig:model} (a), without the steps of pretraining or training the motion encoder $\mathcal{E}$ and decoder $\mathcal{D}$ as autoencoders. Although it maintains the same number of model parameters, it directly learns style transfer on poses, allowing us to assess the impact of our proposed latent stylization.}

% \paragraph{}
\begin{figure*}[t]
	\centering
	\includegraphics[width=\linewidth]{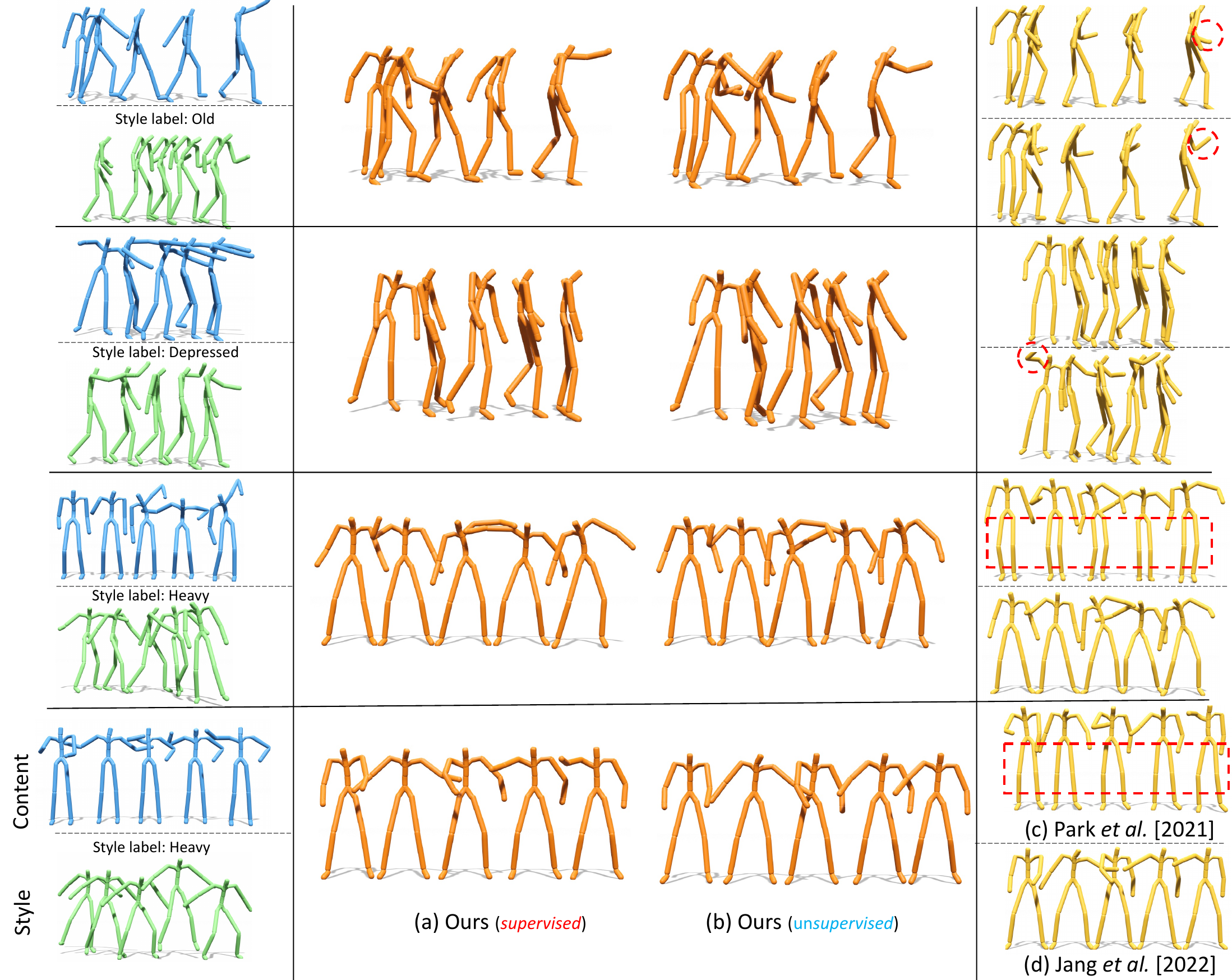}
 	 \vspace{-1.5em}
	\caption{\footnotesize{Qualitative comparisons of motion-based stylization. Given the style motion (green) and content motion (blue), we apply stylization using our methods (orange), ~\citep{park2021diverse} (supervised), and ~\citep{jang2022motion} (unsupervised). The content motions in top two cases come from ~\citep{aberman2020unpaired}, while the bottom two from CMU Mocap~\citep{cmu2021mocap} test sets. Example artifacts 
 are highlighted using red signs. More results are provided in supplementary videos. }}
	\label{fig:qualicomp}
	 \vspace{-1.5em}
\end{figure*}

Overall, our proposed approach consistently achieves appealing performance on a variety of applications across three datasets. In supervised setting, GAN approaches, such as ~\citep{aberman2020unpaired} and ~\citep{park2021diverse}, tend to overfit on one dataset and find difficult on scaling to other motions. For example, ~\citep{park2021diverse} earns the highest achievement on \textit{style recognition} on \citep{aberman2020unpaired}, as 97.1\%, while underperforms on the other two unseen datasets, with style accuracy of 81.3\% and 79.6\%. Furthermore, these methods usually fall short in preserving content, as evidenced by the low content accuracy (31.8\% and 44.1\%) in \cref{tab:xia}. ~\citep{jang2022motion} is shown to be a strong unsupervised baseline; it gains comparable and robust performance on different datasets, which though still suffers from content preservation. On the contrary, our supervised and unsupervised models commonly maintain high style accuracy over 90\% and 80\% respectively, with minimal loss on content semantics. Among all variants, \textit{latent} stylization improves the performance on almost all aspects, including generalization ability, with slight compromise on diversity. \textit{Ours (V)} tends to own higher success rate of style transfer, while \textit{ours (A)} typically outperforms on maintaining content (\textit{i.e.,} \textit{Geo Dis} and \textit{Content Accuracy}).

\paragraph{User Study.} 
In addition, an user study on Amazon Mechanical Turk is conducted to perceptually evaluate our motion stylization results. 50 comparison pairs (on CMU Mocap~\citep{cmu2021mocap}) between each baseline model and our approach, in the corresponding setting, are generated and shown to 4 users, who are asked to choose their favored one regarding realism and stylization quality. Overall, we collect 992 responses from 27 AMT users who have \textit{master} recognition. As shown in \Cref{tab:userstudy}, our method earns more user appreciation over most of the baselines by a large margin. Further user study details are provided in~\Cref{sec:user_study}.%In the unsupervised setting, our method also slightly outperforms the state-of-the-art, ~\citep{jang2022motion}, which consists of heavily designed multi-scale skeleton-based GCN networks for body part style transfer. 

\paragraph{Efficiency.} \Cref{tab:infertime} presents the comparisons of average time cost for a single forward pass with 160-frame motion inputs, evaluated on a single Tesla P100 16G GPU. Previous methods apply style injection at each generator layers until the motion output, and usually involves  computationally intensive operations such as multi-scale skeleton-based GCN and forward-loop kinematics. Benefiting from our latent stylization and lightweight network design, our model appears to be much faster and shows the potential for real time applications.

% \begin{figure*}[t]
% 	\centering
% 	\includegraphics[width=\linewidth]{figures/User_Study_expertmix11.pdf}
%  	 \vspace{-1.5em}
% 	\caption{\textbf{User study.} \yxmu{Place in the supplementary doc?}}
% 	\label{fig:userstudy}
% 	 % \vspace{-1.5em}
% \end{figure*}

\vspace{-0.5em}
\subsection{Qualitative Results}
\label{subsec:qualitative_results}
\Cref{fig:qualicomp} presents the visual comparison results on test sets of ~\citep{aberman2020unpaired} (top two) and CMU Mocap~\citep{cmu2021mocap} (bottom two), in supervised (ours vs.~\citep{park2021diverse}) and unsupervised (ours vs. ~\citep{jang2022motion}) settings. For our model, we use \textit{ours(V)} by default. In unsupervised setting, ~\citep{jang2022motion} has comparable performance on transferring style from style motion to content motion; but it sometimes changes the actions from content motion, as indicated in red circles. Supervised baseline~\citep{park2021diverse} follows similar trend. Moreover, the results of ~\citep{park2021diverse} on CMU Mocap~\citep{aberman2020unpaired} commonly fail to capture the style information from input style motion. This also agrees with the observation of limited generalization ability of GAN-based models in~\cref{tab:aberman_cmu}. Other artifacts such as unnatural poses (~\citep{aberman2020unpaired,park2021diverse}) and foot sliding(~\citep{jang2022motion}) can be better viewed in the supplementary video. This can be partially attributed to the baselines directly applying global velocities of content motion for stylization results. In contrast, our approach show reliable performance on both maintaining content semantics and capturing style characteristics for robust stylization.

\begin{figure*}[t]
	\centering
	\includegraphics[width=0.9\linewidth]{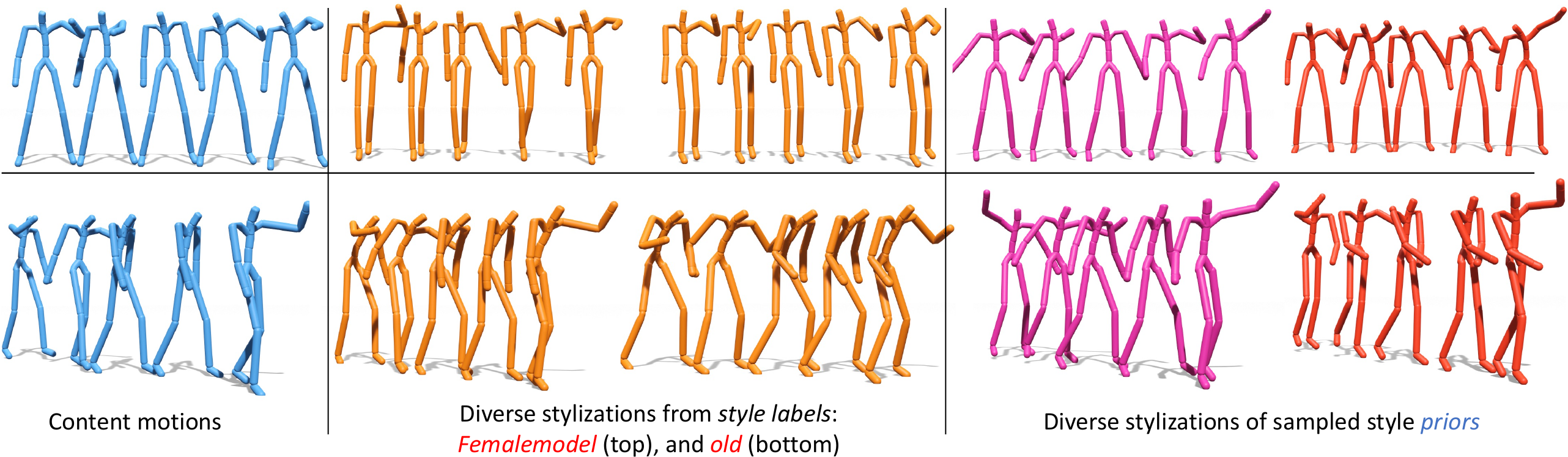}
 	 \vspace{-1em}
	\caption{\footnotesize{Two examples of diverse label-based stylization (middle) and prior-based stylization (right).} }
	\label{fig:novel_and_diverse}
	 \vspace{-0.5em}
\end{figure*}
\begin{figure*}[t]
	\centering
	\includegraphics[width=0.9\linewidth]{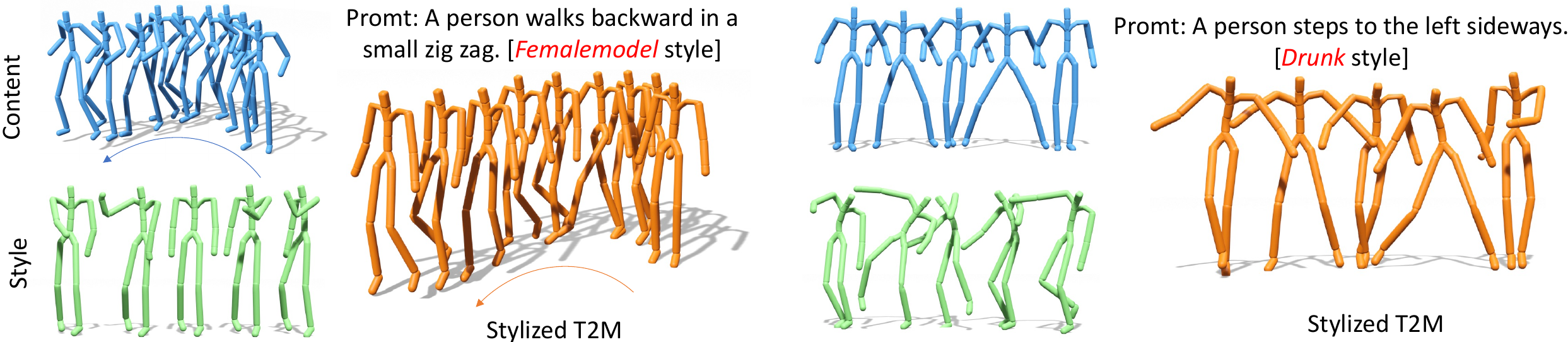}
 	 % \vspace{-1.5em}
	\caption{\footnotesize{Two stylized text2motion examples, by applying our method behind text2motion~\citep{guo2022generating}.}}
	\label{fig:t2m}
	 \vspace{-1.3em}
\end{figure*}

\paragraph{Diverse and Stochastic Stylization.} Our approach allows for diverse label-based and stochastic prior-based stylization. As presented in ~\Cref{fig:novel_and_diverse}, for label-based stylization, taken one content motion and style label as input, our model is able to generate multiple stylized results with inner-class variations, \textit{i.e.,} different manners of \textit{old} man walking. On the other hand, our prior-based stylization can produce results with very distinct styles that are learned unsupervisedly from motion data. These styles are undefined, and possibly non-existing in training data.

\paragraph{Stylized Text2Motion.} Text-to-motion synthesis has attracted significant interests in recent years~\citep{guo2022generating, guo2022tm2t, petrovich2022temos, tevet2022motionclip}; however, the results often exhibit limited style expression. Here in ~\cref{fig:t2m}, we demonstrate the feasibility of generating stylistic human motions from the text prompt, by simply plugging our model behind an text2motion generator~\citep{guo2022generating}. It is worth noting that the motions from~\citep{guo2022generating} differs greatly from our learning data in terms of motion domain and frame rate (20 fps vs. ours 30 fps).    

\begin{wrapfigure}{r}{0.4\textwidth}
  % \centering
       	 \vspace{-1.5em}

    \includegraphics[width=\linewidth]{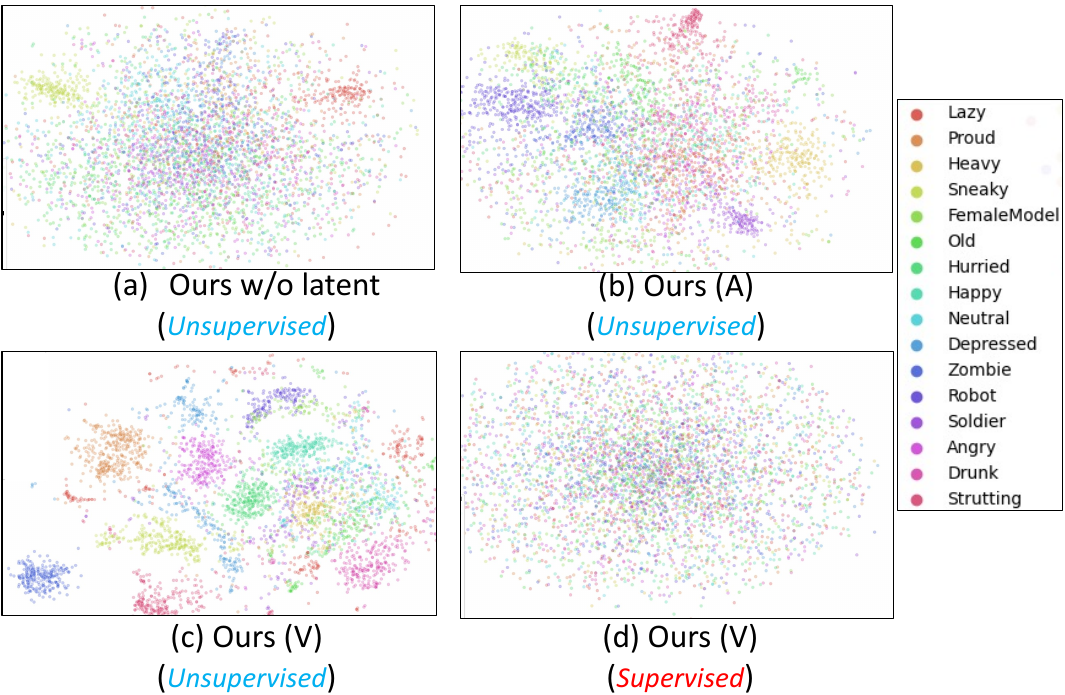}
     	 \vspace{-1.5em}
	\caption{\footnotesize{Style code visualization.} }
	\label{fig:latent_vis}
  	 \vspace{-1em}
 \end{wrapfigure}

\paragraph{Style Code Visualization.} 

Figure \ref{fig:latent_vis} displays the t-SNE 2D projection of our extracted style codes using four model variants, where each sample is color-coded according to its label. In the unsupervised setting, each style code is associated with global style features that are expected to be distinctive with respect to the style category. It can be observed that our \textit{latent stylization} method produces clearer style clusters aligned with style labels compared to our non-latent method, with VAE-based latent model (\textit{ours(V)}) performing the best. While in supervised setting, as discussed in~\cref{subsubsec:learning_scheme}, our approach learns label-invariant style features (\cref{fig:latent_vis} (d)); These style features may arise from individual and environmental factors.

Due to the space limit, we relegate interpolation results (\cref{sec:interpolation}), component analysis (\cref{sec:user_study}), limitations discussion (\cref{sec:failure_cases}), and more visualizations to supplementary files.

\section{Conclusion}
\label{sec:conclusion}
Our work looks into the problem of 3D human motion stylization, with particular emphasis on generative stylization in the neural latent space. Our approach learns a probabilistic style space from motion latent codes; this space allows style sampling for stylization conditioned on reference style motion, target style label, or free-form novel re-stylization. Experiments on three mocap datasets also demonstrate other merits of our model such as better generalization ability, flexibility in style controls, stylization diversity  and efficiency in forward pass.

\paragraph{Ethics Statement.} In practice use, our method is likely to cause demographic discrimination, as it involves stereotypical styles related to gender (\textit{Femalemodel}), age (\textit{old}) and occupation (\textit{soldier}).

\paragraph{Reproducibility Statement.} We have made our best efforts to ensure reproducibility, including but not limited to: 1) detailed description of our implementation details in~\Cref{sec:implementation}; 2) detailed description of our baseline implementation in~\Cref{subsec:baseline}; 3) graphic illustration of our model architecture in~\Cref{fig:fullarc,fig:mvae}; and 4) uploaded codes as supplementary files.

\bibliography{iclr2024_conference}

\begin{thebibliography}{43}
\providecommand{\natexlab}[1]{#1}
\providecommand{\url}[1]{\texttt{#1}}
\expandafter\ifx\csname urlstyle\endcsname\relax
  \providecommand{\doi}[1]{doi: #1}\else
  \providecommand{\doi}{doi: \begingroup \urlstyle{rm}\Url}\fi

\bibitem[Aberman et~al.(2020)Aberman, Weng, Lischinski, Cohen-Or, and Chen]{aberman2020unpaired}
Kfir Aberman, Yijia Weng, Dani Lischinski, Daniel Cohen-Or, and Baoquan Chen.
\newblock Unpaired motion style transfer from video to animation.
\newblock \emph{ACM Transactions on Graphics (TOG)}, 39\penalty0 (4):\penalty0 64--1, 2020.

\bibitem[Chen et~al.(2022)Chen, Jiang, Liu, Huang, Fu, Chen, Yu, and Yu]{chen2022executing}
Xin Chen, Biao Jiang, Wen Liu, Zilong Huang, Bin Fu, Tao Chen, Jingyi Yu, and Gang Yu.
\newblock Executing your commands via motion diffusion in latent space.
\newblock \emph{arXiv preprint arXiv:2212.04048}, 2022.

\bibitem[CMU()]{cmu2021mocap}
CMU.
\newblock Carnegie-mellon mocap database.
\newblock Retrieved from \url{http://mocap.cs.cmu.edu}.

\bibitem[Ding et~al.(2021)Ding, Yang, Hong, Zheng, Zhou, Yin, Lin, Zou, Shao, Yang, et~al.]{ding2021cogview}
Ming Ding, Zhuoyi Yang, Wenyi Hong, Wendi Zheng, Chang Zhou, Da~Yin, Junyang Lin, Xu~Zou, Zhou Shao, Hongxia Yang, et~al.
\newblock Cogview: Mastering text-to-image generation via transformers.
\newblock \emph{Advances in Neural Information Processing Systems}, 34:\penalty0 19822--19835, 2021.

\bibitem[Du et~al.(2019)Du, Herrmann, Sprenger, Fischer, and Slusallek]{du2019stylistic}
Han Du, Erik Herrmann, Janis Sprenger, Klaus Fischer, and Philipp Slusallek.
\newblock Stylistic locomotion modeling and synthesis using variational generative models.
\newblock In \emph{Proceedings of the 12th ACM SIGGRAPH Conference on Motion, Interaction and Games}, pp.\  1--10, 2019.

\bibitem[Esser et~al.(2021{\natexlab{a}})Esser, Rombach, Blattmann, and Ommer]{esser2021imagebart}
Patrick Esser, Robin Rombach, Andreas Blattmann, and Bjorn Ommer.
\newblock Imagebart: Bidirectional context with multinomial diffusion for autoregressive image synthesis.
\newblock \emph{Advances in Neural Information Processing Systems}, 34:\penalty0 3518--3532, 2021{\natexlab{a}}.

\bibitem[Esser et~al.(2021{\natexlab{b}})Esser, Rombach, and Ommer]{esser2021taming}
Patrick Esser, Robin Rombach, and Bjorn Ommer.
\newblock Taming transformers for high-resolution image synthesis.
\newblock In \emph{Proceedings of the IEEE/CVF conference on computer vision and pattern recognition}, pp.\  12873--12883, 2021{\natexlab{b}}.

\bibitem[Fu et~al.(2022)Fu, Zhan, Chen, Ritchie, and Sridhar]{fu2022shapecrafter}
Rao Fu, Xiao Zhan, Yiwen Chen, Daniel Ritchie, and Srinath Sridhar.
\newblock Shapecrafter: A recursive text-conditioned 3d shape generation model.
\newblock \emph{arXiv preprint arXiv:2207.09446}, 2022.

\bibitem[Gatys et~al.(2016)Gatys, Ecker, and Bethge]{gatys2016image}
Leon~A Gatys, Alexander~S Ecker, and Matthias Bethge.
\newblock Image style transfer using convolutional neural networks.
\newblock In \emph{Proceedings of the IEEE conference on computer vision and pattern recognition}, pp.\  2414--2423, 2016.

\bibitem[Gong et~al.(2023)Gong, Lian, Chang, Guo, Zuo, Jiang, and Wang]{gong2023tm2d}
Kehong Gong, Dongze Lian, Heng Chang, Chuan Guo, Xinxin Zuo, Zihang Jiang, and Xinchao Wang.
\newblock Tm2d: Bimodality driven 3d dance generation via music-text integration.
\newblock \emph{arXiv preprint arXiv:2304.02419}, 2023.

\bibitem[Guo et~al.(2022{\natexlab{a}})Guo, Zou, Zuo, Wang, Ji, Li, and Cheng]{guo2022generating}
Chuan Guo, Shihao Zou, Xinxin Zuo, Sen Wang, Wei Ji, Xingyu Li, and Li~Cheng.
\newblock Generating diverse and natural 3d human motions from text.
\newblock In \emph{Proceedings of the IEEE/CVF Conference on Computer Vision and Pattern Recognition}, pp.\  5152--5161, 2022{\natexlab{a}}.

\bibitem[Guo et~al.(2022{\natexlab{b}})Guo, Zuo, Wang, and Cheng]{guo2022tm2t}
Chuan Guo, Xinxin Zuo, Sen Wang, and Li~Cheng.
\newblock Tm2t: Stochastic and tokenized modeling for the reciprocal generation of 3d human motions and texts.
\newblock In \emph{Computer Vision--ECCV 2022: 17th European Conference, Tel Aviv, Israel, October 23--27, 2022, Proceedings, Part XXXV}, pp.\  580--597. Springer, 2022{\natexlab{b}}.

\bibitem[Holden et~al.(2016)Holden, Saito, and Komura]{holden2016deep}
Daniel Holden, Jun Saito, and Taku Komura.
\newblock A deep learning framework for character motion synthesis and editing.
\newblock \emph{ACM Transactions on Graphics (TOG)}, 35\penalty0 (4):\penalty0 1--11, 2016.

\bibitem[Holden et~al.(2017)Holden, Habibie, Kusajima, and Komura]{holden2017fast}
Daniel Holden, Ikhsanul Habibie, Ikuo Kusajima, and Taku Komura.
\newblock Fast neural style transfer for motion data.
\newblock \emph{IEEE computer graphics and applications}, 37\penalty0 (4):\penalty0 42--49, 2017.

\bibitem[Hu et~al.(2023)Hu, Chen, and Luo]{hu2023lamd}
Yaosi Hu, Zhenzhong Chen, and Chong Luo.
\newblock Lamd: Latent motion diffusion for video generation.
\newblock \emph{arXiv preprint arXiv:2304.11603}, 2023.

\bibitem[Huang \& Belongie(2017)Huang and Belongie]{huang2017arbitrary}
Xun Huang and Serge Belongie.
\newblock Arbitrary style transfer in real-time with adaptive instance normalization.
\newblock In \emph{Proceedings of the IEEE international conference on computer vision}, pp.\  1501--1510, 2017.

\bibitem[Isola et~al.(2017)Isola, Zhu, Zhou, and Efros]{isola2017image}
Phillip Isola, Jun-Yan Zhu, Tinghui Zhou, and Alexei~A Efros.
\newblock Image-to-image translation with conditional adversarial networks.
\newblock In \emph{Proceedings of the IEEE conference on computer vision and pattern recognition}, pp.\  1125--1134, 2017.

\bibitem[Jang et~al.(2022)Jang, Park, and Lee]{jang2022motion}
Deok-Kyeong Jang, Soomin Park, and Sung-Hee Lee.
\newblock Motion puzzle: Arbitrary motion style transfer by body part.
\newblock \emph{ACM Transactions on Graphics (TOG)}, 41\penalty0 (3):\penalty0 1--16, 2022.

\bibitem[Johnson et~al.(2016)Johnson, Alahi, and Fei-Fei]{johnson2016perceptual}
Justin Johnson, Alexandre Alahi, and Li~Fei-Fei.
\newblock Perceptual losses for real-time style transfer and super-resolution.
\newblock In \emph{Computer Vision--ECCV 2016: 14th European Conference, Amsterdam, The Netherlands, October 11-14, 2016, Proceedings, Part II 14}, pp.\  694--711. Springer, 2016.

\bibitem[Karras et~al.(2019)Karras, Laine, and Aila]{karras2019style}
Tero Karras, Samuli Laine, and Timo Aila.
\newblock A style-based generator architecture for generative adversarial networks.
\newblock In \emph{Proceedings of the IEEE/CVF conference on computer vision and pattern recognition}, pp.\  4401--4410, 2019.

\bibitem[Kingma \& Welling(2013)Kingma and Welling]{kingma2013auto}
Diederik~P Kingma and Max Welling.
\newblock Auto-encoding variational bayes.
\newblock \emph{arXiv preprint arXiv:1312.6114}, 2013.

\bibitem[LeCun et~al.(1995)LeCun, Bengio, et~al.]{lecun1995convolutional}
Yann LeCun, Yoshua Bengio, et~al.
\newblock Convolutional networks for images, speech, and time series.
\newblock \emph{The handbook of brain theory and neural networks}, 3361\penalty0 (10):\penalty0 1995, 1995.

\bibitem[Lee et~al.(2019)Lee, Yang, Liu, Wang, Lu, Yang, and Kautz]{lee2019dancing}
Hsin-Ying Lee, Xiaodong Yang, Ming-Yu Liu, Ting-Chun Wang, Yu-Ding Lu, Ming-Hsuan Yang, and Jan Kautz.
\newblock Dancing to music.
\newblock \emph{Advances in neural information processing systems}, 32, 2019.

\bibitem[Park et~al.(2021)Park, Jang, and Lee]{park2021diverse}
Soomin Park, Deok-Kyeong Jang, and Sung-Hee Lee.
\newblock Diverse motion stylization for multiple style domains via spatial-temporal graph-based generative model.
\newblock \emph{Proceedings of the ACM on Computer Graphics and Interactive Techniques}, 4\penalty0 (3):\penalty0 1--17, 2021.

\bibitem[Park et~al.(2020)Park, Zhu, Wang, Lu, Shechtman, Efros, and Zhang]{park2020swapping}
Taesung Park, Jun-Yan Zhu, Oliver Wang, Jingwan Lu, Eli Shechtman, Alexei Efros, and Richard Zhang.
\newblock Swapping autoencoder for deep image manipulation.
\newblock \emph{Advances in Neural Information Processing Systems}, 33:\penalty0 7198--7211, 2020.

\bibitem[Petrovich et~al.(2022)Petrovich, Black, and Varol]{petrovich2022temos}
Mathis Petrovich, Michael~J Black, and G{\"u}l Varol.
\newblock Temos: Generating diverse human motions from textual descriptions.
\newblock In \emph{Computer Vision--ECCV 2022: 17th European Conference, Tel Aviv, Israel, October 23--27, 2022, Proceedings, Part XXII}, pp.\  480--497. Springer, 2022.

\bibitem[Raab et~al.(2023)Raab, Leibovitch, Tevet, Arar, Bermano, and Cohen-Or]{raab2023single}
Sigal Raab, Inbal Leibovitch, Guy Tevet, Moab Arar, Amit~H Bermano, and Daniel Cohen-Or.
\newblock Single motion diffusion.
\newblock \emph{arXiv preprint arXiv:2302.05905}, 2023.

\bibitem[Radford et~al.(2021)Radford, Kim, Hallacy, Ramesh, Goh, Agarwal, Sastry, Askell, Mishkin, Clark, et~al.]{radford2021learning}
Alec Radford, Jong~Wook Kim, Chris Hallacy, Aditya Ramesh, Gabriel Goh, Sandhini Agarwal, Girish Sastry, Amanda Askell, Pamela Mishkin, Jack Clark, et~al.
\newblock Learning transferable visual models from natural language supervision.
\newblock In \emph{International conference on machine learning}, pp.\  8748--8763. PMLR, 2021.

\bibitem[Ramesh et~al.(2022)Ramesh, Dhariwal, Nichol, Chu, and Chen]{ramesh2022hierarchical}
Aditya Ramesh, Prafulla Dhariwal, Alex Nichol, Casey Chu, and Mark Chen.
\newblock Hierarchical text-conditional image generation with clip latents.
\newblock \emph{arXiv preprint arXiv:2204.06125}, 2022.

\bibitem[Rombach et~al.(2022)Rombach, Blattmann, Lorenz, Esser, and Ommer]{rombach2022high}
Robin Rombach, Andreas Blattmann, Dominik Lorenz, Patrick Esser, and Bj{\"o}rn Ommer.
\newblock High-resolution image synthesis with latent diffusion models.
\newblock In \emph{Proceedings of the IEEE/CVF Conference on Computer Vision and Pattern Recognition}, pp.\  10684--10695, 2022.

\bibitem[Smith et~al.(2019)Smith, Cao, Neff, and Wang]{smith_efficient_2019}
Harrison~Jesse Smith, Chen Cao, Michael Neff, and Yingying Wang.
\newblock Efficient {Neural} {Networks} for {Real}-time {Motion} {Style} {Transfer}.
\newblock \emph{Proceedings of the ACM on Computer Graphics and Interactive Techniques}, 2\penalty0 (2):\penalty0 1--17, July 2019.
\newblock ISSN 2577-6193.
\newblock \doi{10.1145/3340254}.
\newblock URL \url{https://dl.acm.org/doi/10.1145/3340254}.

\bibitem[Tao et~al.(2022)Tao, Zhan, Chen, and van~de Panne]{tao_style-erd_2022}
Tianxin Tao, Xiaohang Zhan, Zhongquan Chen, and Michiel van~de Panne.
\newblock Style-{ERD}: {Responsive} and {Coherent} {Online} {Motion} {Style} {Transfer}.
\newblock In \emph{2022 {IEEE}/{CVF} {Conference} on {Computer} {Vision} and {Pattern} {Recognition} ({CVPR})}, pp.\  6583--6593, New Orleans, LA, USA, June 2022. IEEE.
\newblock ISBN 978-1-66546-946-3.
\newblock \doi{10.1109/CVPR52688.2022.00648}.
\newblock URL \url{https://ieeexplore.ieee.org/document/9879697/}.

\bibitem[Tevet et~al.(2022)Tevet, Gordon, Hertz, Bermano, and Cohen-Or]{tevet2022motionclip}
Guy Tevet, Brian Gordon, Amir Hertz, Amit~H Bermano, and Daniel Cohen-Or.
\newblock Motionclip: Exposing human motion generation to clip space.
\newblock In \emph{Computer Vision--ECCV 2022: 17th European Conference, Tel Aviv, Israel, October 23--27, 2022, Proceedings, Part XXII}, pp.\  358--374. Springer, 2022.

\bibitem[Ulyanov et~al.(2016{\natexlab{a}})Ulyanov, Lebedev, Vedaldi, and Lempitsky]{ulyanov2016texture}
Dmitry Ulyanov, Vadim Lebedev, Andrea Vedaldi, and Victor Lempitsky.
\newblock Texture networks: Feed-forward synthesis of textures and stylized images.
\newblock \emph{arXiv preprint arXiv:1603.03417}, 2016{\natexlab{a}}.

\bibitem[Ulyanov et~al.(2016{\natexlab{b}})Ulyanov, Vedaldi, and Lempitsky]{ulyanov2016instance}
Dmitry Ulyanov, Andrea Vedaldi, and Victor Lempitsky.
\newblock Instance normalization: The missing ingredient for fast stylization.
\newblock \emph{arXiv preprint arXiv:1607.08022}, 2016{\natexlab{b}}.

\bibitem[Van Den~Oord et~al.(2017)Van Den~Oord, Vinyals, et~al.]{van2017neural}
Aaron Van Den~Oord, Oriol Vinyals, et~al.
\newblock Neural discrete representation learning.
\newblock \emph{Advances in neural information processing systems}, 30, 2017.

\bibitem[Wen et~al.(2021)Wen, Yang, Fu, Gao, Sun, and Liu]{wen2021autoregressive}
Yu-Hui Wen, Zhipeng Yang, Hongbo Fu, Lin Gao, Yanan Sun, and Yong-Jin Liu.
\newblock Autoregressive stylized motion synthesis with generative flow.
\newblock In \emph{Proceedings of the IEEE/CVF Conference on Computer Vision and Pattern Recognition}, pp.\  13612--13621, 2021.

\bibitem[Xia et~al.(2015)Xia, Wang, Chai, and Hodgins]{xia2015realtime}
Shihong Xia, Congyi Wang, Jinxiang Chai, and Jessica Hodgins.
\newblock Realtime style transfer for unlabeled heterogeneous human motion.
\newblock \emph{ACM Transactions on Graphics (TOG)}, 34\penalty0 (4):\penalty0 1--10, 2015.

\bibitem[Yan et~al.(2021)Yan, Zhang, Abbeel, and Srinivas]{yan2021videogpt}
Wilson Yan, Yunzhi Zhang, Pieter Abbeel, and Aravind Srinivas.
\newblock Videogpt: Video generation using vq-vae and transformers.
\newblock \emph{arXiv preprint arXiv:2104.10157}, 2021.

\bibitem[Yumer \& Mitra(2016)Yumer and Mitra]{yumer_spectral_2016}
M.~Ersin Yumer and Niloy~J. Mitra.
\newblock Spectral style transfer for human motion between independent actions.
\newblock \emph{ACM Transactions on Graphics}, 35\penalty0 (4):\penalty0 1--8, July 2016.
\newblock ISSN 0730-0301, 1557-7368.
\newblock \doi{10.1145/2897824.2925955}.
\newblock URL \url{https://dl.acm.org/doi/10.1145/2897824.2925955}.

\bibitem[Zeng et~al.(2022)Zeng, Vahdat, Williams, Gojcic, Litany, Fidler, and Kreis]{zeng2022lion}
Xiaohui Zeng, Arash Vahdat, Francis Williams, Zan Gojcic, Or~Litany, Sanja Fidler, and Karsten Kreis.
\newblock Lion: Latent point diffusion models for 3d shape generation.
\newblock In \emph{Advances in Neural Information Processing Systems (NeurIPS)}, 2022.

\bibitem[Zhou et~al.(2019)Zhou, Barnes, Lu, Yang, and Li]{zhou2019continuity}
Yi~Zhou, Connelly Barnes, Jingwan Lu, Jimei Yang, and Hao Li.
\newblock On the continuity of rotation representations in neural networks.
\newblock In \emph{Proceedings of the IEEE/CVF Conference on Computer Vision and Pattern Recognition}, pp.\  5745--5753, 2019.

\bibitem[Zhu et~al.(2017)Zhu, Park, Isola, and Efros]{zhu2017unpaired}
Jun-Yan Zhu, Taesung Park, Phillip Isola, and Alexei~A Efros.
\newblock Unpaired image-to-image translation using cycle-consistent adversarial networks.
\newblock In \emph{Proceedings of the IEEE international conference on computer vision}, pp.\  2223--2232, 2017.

\end{thebibliography}
\bibliographystyle{iclr2024_conference}

\newpage
\appendix
This supplementary file provides additional ablation analysis of our design components (\cref{subsec:ablation}) and weight of various loss terms (\cref{sec:experiment}), content and feature visualization(\cref{subsec:latent_visualization}), implementation details (\cref{sec:implementation}), Evaluation Metric (\cref{sec:metric}), cross-style and homo-style interpolation results (\cref{sec:interpolation}), user study introduction (\cref{sec:user_study}) and failure cases (\cref{sec:failure_cases}).

\paragraph{Video.} We also provide several supplementary videos, which contains dynamic animations of our stylization results, visual comparisons, interpolation, stylized text2motion and failure cases. We strongly encourage our audience to watch these videos. It will be much helpful to understand our work. The videos are submitted along with supplementary files, or accessible online (1080P): \href{https://drive.google.com/drive/folders/1UeGuE1qCceLFQJa3vpYoOHC2MoLdBifK?usp=sharing}{https://drive.google.com/drive/folders/1UeGuE1qCceLFQJa3vpYoOHC2MoLdBifK?usp=sharing}

\paragraph{Code and Model.} The code of our approach and implemented baselines are also submitted for reference. Code and trained model will be publicly available upon acceptance.

\revise{\section{Ablation Analysis}}
\label{subsec:ablation}
\begin{table}[ht]
    % \centering
    \scalebox{0.73}{
    \begin{tabular}{l l c c c c c c c c c c}
    \toprule
    % \multirow{2}{*}{Methods} & \multicolumn{4}{c}{\DN (Coarse-grained)} & & \multicolumn{4}{c}{HumanAct(Fine-grained)} \\
    % \cline{2-5}
    % \cline{7-10}
    %                 & FID$\downarrow$ & Accuracy$\uparrow$ & Diversity$\rightarrow$& MModality$\rightarrow$ &  & FID$\downarrow$ & Accuracy$\uparrow$ & Diversity$\rightarrow$ & MModality$\rightarrow$\\
    \multirow{2}{*}{S / U}  & \multirow{2}{*}{Method}  & \multicolumn{3}{c}{~\citep{aberman2020unpaired}} & ~ &\multicolumn{4}{c}{~\citep{xia2015realtime}} \\
    \cline{3-5}
    \cline{7-10}
    ~& ~& Style Acc$\uparrow$ & Style FID $\downarrow$ & Geo Dis$\downarrow$ & ~& Style Acc$\uparrow$ & Content Acc$\uparrow$ & Content FID$\downarrow$ & Geo Dis $\downarrow$  \\
    \midrule
    S & Ours (A) & \etb{0.945}{007} & \etb{0.020}{002} & \et{\textbf{0.344}}{002} & & \et{\textbf{0.926}}{008} & \et{\textbf{0.674}}{011} & \et{\textbf{0.189}}{005} & \et{\textbf{0.680}}{002} \\
    \midrule
    ~ & \textit{w/o} latent &  \ets{0.932}{008} & \et{0.022}{002} & \et{0.463}{003} & & \et{0.851}{012} & \ets{0.654}{012} & \et{0.258}{007} & \et{0.707}{003} \\
    ~ & \textit{w/o} prob-style & \et{0.913}{007} & \et{0.022}{002} & \et{0.509}{004} & & \et{0.870}{010} & \et{0.524}{015} & \et{0.249}{008} & \et{0.767}{004} \\
    ~ & \textit{w/o} homo-style & \et{0.883}{012} & \et{0.032}{004} & \et{0.507}{003} & & \et{0.851}{012} & \et{0.537}{016} & \et{0.232}{006} & \et{0.760}{004} \\
    ~ & \textit{w/o} autoencoding & \et{0.900}{010} & \et{0.026}{002} & \et{0.427}{003} & & \ets{0.879}{010} & \et{0.634}{011} & \ets{0.198}{005} & \et{0.720}{004} \\
    ~ & \textit{w/o} cycle-recon & \et{0.917}{009} & \ets{0.021}{002} & \ets{0.385}{003} & & \et{0.872}{006} & \et{0.627}{011} & \et{0.208}{004} & \ets{0.699}{002} \\
    \midrule
    U & Ours (A) & \etb{0.804}{011} & \etb{0.040}{003} & \et{\textbf{0.441}}{003} & & \etb{0.814}{011} & \ets{0.588}{010} & \etb{0.217}{006} & \et{0.735}{003} \\
    \midrule
    ~ & \textit{w/o} latent &  \ets{0.780}{014} & \et{0.048}{003} & \et{0.466}{004} & & \et{0.734}{014} & \et{0.584}{011} & \et{0.272}{008} & \ets{0.721}{003} \\
    ~ & \textit{w/o} prob-style & \et{0.734}{018} & \et{0.058}{004} & \et{0.461}{003} & & \et{0.666}{016} & \etb{0.597}{015} & \et{0.270}{010} & \etb{0.718}{003} \\
    ~ & \textit{w/o} homo-style & \et{0.753}{016} & \et{0.050}{002} & \et{0.513}{003} & & \et{0.730}{009} & \et{0.526}{013} & \et{0.250}{005} & \et{0.803}{002} \\
    ~ & \textit{w/o} autoencoding & \et{0.777}{012} & \et{0.049}{004} & \et{0.493}{004} & & \ets{0.811}{011} & \et{0.491}{015} & \ets{0.230}{007} & \et{0.759}{005} \\
    ~ & \textit{w/o} cycle-recon & \et{0.765}{011} & \ets{0.043}{004} & \ets{0.560}{005} & & \et{0.756}{017} & \et{0.479}{013} & \et{0.233}{007} & \et{0.869}{002} \\
    \bottomrule
    \end{tabular}
    }
    \caption{\revise{\footnotesize{Ablation study on \textbf{different components of our model design}. $\pm$ indicates 95\% confidence interval. \textbf{Bold} face indicates the best result, while \underline{underscore} refers to the second best. (S) and (U) denote \textit{supervised} and \textit{unsupervised} setting. Motion-based stylization is presented for both settings. \textit{Prob-style} refers to probabilistic style space.}}}
    \label{tab:ablation_all}
    	 % \vspace{-1em}

\end{table}
\revise{\Cref{tab:ablation_all} presents the results of ablation experiments investigating various components of our latent stylization models. These components include stylization on the latent space (\textit{latent}), the use of a probabilistic style space (\textit{prob-style}), homo-style alignment (\textit{homo-style}), autoencoding, and cycle reconstruction. The experiments are conducted within the framework of Ours (A) and are focused on the task of motion-based stylization. Results are reported on two datasets~\citep{aberman2020unpaired} and~\citep{xia2015realtime}. It's important to note that the dataset of~\citep{xia2015realtime} is exclusively used for testing the generalization ability of our models and has not been used during training.}

\revise{Overall, we observe a notable performance improvement by incorporating different modules into our framework. For instance, our key designs—latent stylization and the use of a probabilistic style space—significantly enhance performance on the unseen \citep{xia2015realtime} dataset, resulting in a 7\% increase in stylization accuracy in the supervised setting. Additionally, homo-style alignment, despite its simplicity, provides a substantial performance boost across all metrics. Notably, content accuracy sees a remarkable improvement of 13\% and 6\% in supervised and unsupervised settings, respectively, underscoring the effectiveness of homo-style alignment in preserving semantic information.}

\revise{In the subsequent sections, we delve into a detailed discussion of three other critical choices in our model architecture and learning scheme: probabilistic (or deterministic) space for content and style features, separate (or end-to-end) training of latent extractor and stylization model, and the incorporation of a global motion predictor.}
\begin{table}[ht]
    % \centering
    \scalebox{0.72}{
    \begin{tabular}{c c c c c c c c c c c c}
    \toprule
    % \multirow{2}{*}{Methods} & \multicolumn{4}{c}{\DN (Coarse-grained)} & & \multicolumn{4}{c}{HumanAct(Fine-grained)} \\
    % \cline{2-5}
    % \cline{7-10}
    %                 & FID$\downarrow$ & Accuracy$\uparrow$ & Diversity$\rightarrow$& MModality$\rightarrow$ &  & FID$\downarrow$ & Accuracy$\uparrow$ & Diversity$\rightarrow$ & MModality$\rightarrow$\\
    \multirow{2}{*}{Content Space}  & \multirow{2}{*}{Style Space}  & \multicolumn{3}{c}{~\citep{aberman2020unpaired}} & ~ &\multicolumn{4}{c}{~\citep{xia2015realtime}} \\
    \cline{3-5}
    \cline{7-10}
    ~& ~& Style Acc$\uparrow$ & Style FID $\downarrow$ & Geo Dis$\downarrow$ & ~& Style Acc$\uparrow$ & Content Acc$\uparrow$ & Content FID$\downarrow$ & Geo Dis $\downarrow$  \\
    \midrule
    D & D & \et{0.913}{007} & \et{0.022}{002} & \et{0.509}{004} & & \et{0.870}{010} & \et{0.524}{015} & \et{0.249}{008} & \et{0.767}{004} \\
    D & P & \et{0.945}{007} & \et{0.020}{002} & \et{\textbf{0.344}}{002} & & \et{\textbf{0.926}}{008} & \et{\textbf{0.674}}{011} & \et{\textbf{0.189}}{005} & \et{\textbf{0.680}}{002} \\
    P & P & \etb{0.947}{001} & \etb{0.017}{001} & \et{0.489}{003} & & \et{0.891}{003} & \et{0.417}{012} & \et{0.322}{011} & \et{0.758}{003} \\
    % \midrule
    
    \bottomrule
    \end{tabular}
    }
    \caption{\revise{\footnotesize{Ablation study on the \textbf{choice of probabilistic (P) or deterministic (D) space} for content and style, in supervised setting. $\pm$ indicates 95\% confidence interval. \textbf{Bold} face indicates the best result, while \underline{underscore} refers to the second best. Motion-based stylization is presented.}}}
    \label{tab:probabilistic}
    	 % \vspace{-1em}

\end{table}

\revise{\noindent\textbf{Probabilistic Modeling of Style and Content Spaces.} \Cref{tab:probabilistic} presents a comparison between deterministic and probabilistic modeling approaches for both style and content spaces. In our study, the introduction of a probabilistic style space not only provides remarkable flexibility during inference, enabling diverse stylization and multiple applications, but it also consistently enhances performance and generalization capabilities. An intriguing aspect to explore is the impact of modeling the content space non-deterministically. As highlighted in~\cref{tab:probabilistic}, we observe that a probabilistic content space achieves superior stylization accuracy on in-domain datasets~\citep{aberman2020unpaired}. However, it exhibits sub-optimal generalization performance on out-domain cases~\citep{xia2015realtime}.}

\begin{table}[ht]
    % \centering
    \scalebox{0.76}{
    \begin{tabular}{c c c c c c c c c c c}
    \toprule
    % \multirow{2}{*}{Methods} & \multicolumn{4}{c}{\DN (Coarse-grained)} & & \multicolumn{4}{c}{HumanAct(Fine-grained)} \\
    % \cline{2-5}
    % \cline{7-10}
    %                 & FID$\downarrow$ & Accuracy$\uparrow$ & Diversity$\rightarrow$& MModality$\rightarrow$ &  & FID$\downarrow$ & Accuracy$\uparrow$ & Diversity$\rightarrow$ & MModality$\rightarrow$\\
    \multirow{2}{*}{Training Strategy}  & \multicolumn{3}{c}{~\citep{aberman2020unpaired}} & ~ &\multicolumn{4}{c}{~\citep{xia2015realtime}} \\
    \cline{2-4}
    \cline{6-9}
    ~& Style Acc$\uparrow$ & Style FID $\downarrow$ & Geo Dis$\downarrow$ & ~& Style Acc$\uparrow$ & Content Acc$\uparrow$ & Content FID$\downarrow$ & Geo Dis $\downarrow$  \\
    \midrule
    Separately & \etb{0.945}{007} & \etb{0.020}{002} & \et{\textbf{0.344}}{002} & & \et{\textbf{0.926}}{008} & \et{\textbf{0.674}}{011} & \et{\textbf{0.189}}{005} & \et{\textbf{0.680}}{002} \\
    % \midrule
    End-to-end & \et{0.125}{010} & \et{1.521}{024} & \et{0.577}{001} & & \et{0.174}{014} & \et{0.293}{002} & \et{1.417}{009} & \et{0.700}{001} \\
    
    \bottomrule
    \end{tabular}
    }
    \caption{\revise{\footnotesize{Ablation study on \textbf{separately or end-to-end training} the latent model and stylization model, in supervised setting. $\pm$ indicates 95\% confidence interval. \textbf{Bold} face indicates the best result, while \underline{underscore} refers to the second best. (S) and (U) denote \textit{supervised} and \textit{unsupervised} setting. Motion-based stylization is presented.}}}
    \label{tab:end-to-end}
    	 % \vspace{-1em}

\end{table}

\revise{\noindent\textbf{Separate / End-to-end Training.} Our two-stage framework can alternatively be trained in an end-to-end fashion. We also conduct ablation analysis to evaluate the impact of such choice of training strategy. The results are presented in~\Cref{tab:end-to-end}. In practice, we observed that end-to-end training posed significant challenges. The model struggled to simultaneously learn meaningful latent motion representation and effectively transfer style traits between stages. Experimental results align with this observation, revealing that stylization accuracy is merely around 15\% on both datasets in the end-to-end training scenario, in contrast to the accuracy of 92\% achieved by stage-by-stage training.}

\begin{table}[ht]
    \centering
    \scalebox{0.8}{
    \begin{tabular}{l c c c c c c c  c}
    \toprule
    % \multirow{2}{*}{Methods} & \multicolumn{4}{c}{\DN (Coarse-grained)} & & \multicolumn{4}{c}{HumanAct(Fine-grained)} \\
    % \cline{2-5}
    % \cline{7-10}
    %                 & FID$\downarrow$ & Accuracy$\uparrow$ & Diversity$\rightarrow$& MModality$\rightarrow$ &  & FID$\downarrow$ & Accuracy$\uparrow$ & Diversity$\rightarrow$ & MModality$\rightarrow$\\
        %\multirow{2}{*}{S / U}  & \multirow{2}{*}{Method}  & \multicolumn{3}{c}{~\citep{aberman2020unpaired}} & ~ &\multicolumn{4}{c}{~\citep{xia2015realtime}} \\

    \multirow{2}{*}{Method} & \multicolumn{2}{c}{~\citep{aberman2020unpaired}} & & \multicolumn{2}{c}{CMU Mocap~\citep{cmu2021mocap}} && \multicolumn{2}{c}{\citep{xia2015realtime}}\\
    \cline{2-3}
    \cline{5-6}
    \cline{8-9}
    ~ &Style Acc$\uparrow$ & Foot Skating$\downarrow$ & & Style Acc$\uparrow$ & Foot Skating$\downarrow$ & &Style Acc$\uparrow$  & Foot Skating$\downarrow$  \\
    \midrule
    Ours (S) & \etb{0.945}{007} & \etb{0.130}{001} & & \et{0.918}{007} & \etb{0.140}{001}& & \etb{0.926}{008}&\etb{0.263}{003}\\
    Ours \textit{w/o} GMP (S) &  \et{0.942}{003} & \et{0.141}{001} & &\etb{0.920}{006}& \et{0.160}{001} & &\et{0.882}{008}& \et{0.331}{002}\\
    \midrule
    Ours (U) & \etb{0.840}{010} &  \etb{0.102}{001}& & \etb{0.828}{010}& \etb{0.099}{001} & & \etb{0.860}{010} & \etb{0.179}{002} \\
    Ours \textit{w/o} GMP (U) & \et{0.817}{013} & \et{0.116}{001}& &\et{0.820}{009} &\et{0.122}{001} & & \et{0.777}{018} & \et{0.307}{002} \\
    \bottomrule
    \end{tabular}
    }
    \caption{\revise{\footnotesize{Ablation study on \textbf{global motion prediction} (GMP, see \cref{subsec:gmp}). The symbol $\pm$ indicates the 95\% confidence interval. \textbf{Bold} indicates the best result. (S) and (U) denote \textit{supervised} and \textit{unsupervised} settings, respectively. Results of motion-based stylization are presented. \textit{Foot skating} is measured by the average velocity of foot joints on the XZ-plane during foot contact.}}}
    \label{tab:gmp}
    	 % \vspace{-1em}

\end{table}

\begin{table}[ht]
    \centering
    \scalebox{0.7}{
    \begin{tabular}{c c c}
        \toprule
        ~\citep{aberman2020unpaired} & CMU Mocap~\citep{cmu2021mocap} & ~\citep{xia2015realtime}\\
        \midrule
        46.2 & 48.7 & 57.7\\
        \bottomrule
    \end{tabular}
    }
    \caption{\revise{\footnotesize{\textbf{Mean Square Error of Root Position Prediction.} The metric is measured in millimeters. Note the dataset of~\citep{xia2015realtime} is untouched during the training of the global motion predictor. }}}
    \label{tab:root_error}
\end{table}

\revise{\paragraph{Global Motion Prediction (GMP).} The primary objective of our global motion prediction is to facilitate adaptive pacing for diverse motion contents and styles. As illustrated in~\cref{tab:root_error}, we quantify the mean square error of GMP in predicting root positions across three test sets, measured in millimeters. Notably, even on the previously unseen dataset~\cite{xia2015realtime}, the lightweight GMP performs admirably, with an error of 57.7 mm.}

\revise{To assess the impact of GMP on stylization performance, we compare against a contrast setting (Ours \textit{w/o} GMP), where global motions are directly obtained from the source content input, akin to previous approaches. Additionally, we introduce a \textit{foot skating} metric to gauge foot sliding artifacts, calculated by the average velocity of foot joints on the XZ-plane during foot contact.\Cref{tab:gmp} showcases motion-based results on \citep{aberman2020unpaired,cmu2021mocap,xia2015realtime} test sets. Across all comparisons, our proposed GMP effectively mitigates foot skating issues. Although 2-dimensional global motion features constitute only a small fraction of the entire 260-dimensional pose vectors, it makes considerable difference on the dataset of ~\citep{xia2015realtime}, improving the stylization accuracy by around 9\%.  In our 3rd and 4th supplementary videos, we also illustrate how our GMP enables adaptive pacing in different stylization outcomes (label-based and motion-based) for the same content.}

\section{Feature Visualization}
\label{subsec:latent_visualization}
\begin{figure*}[ht]
	\centering
	\includegraphics[width=\linewidth]{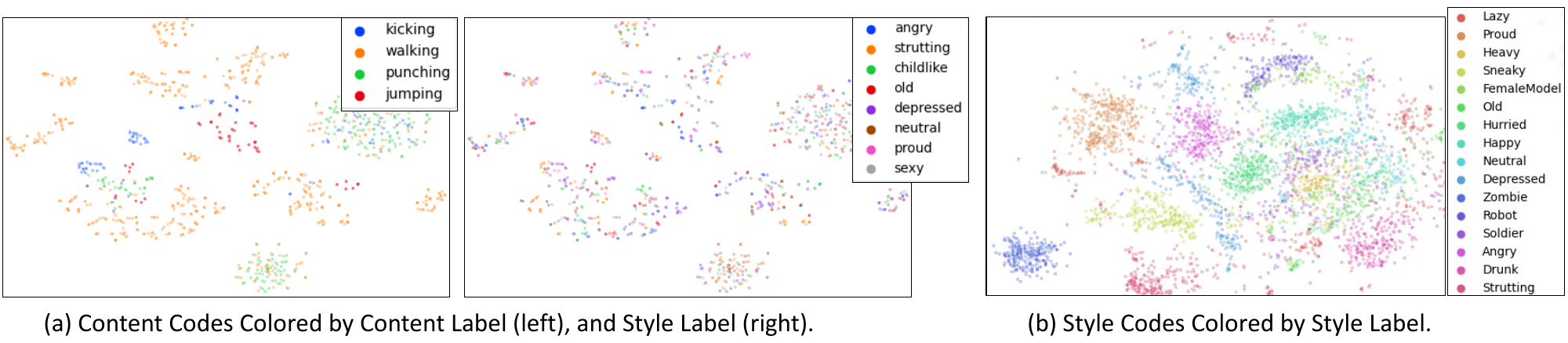}
 	 % \vspace{-1.5em}
	\caption{\revise{\footnotesize{\textbf{Latent Visualization.} Panel (a) displays the projection of the \textbf{identical} set of content codes onto a 2D space using t-SNE, colored according to content labels (left) and style labels (right). This visualization suggests that content codes faithfully capture content traits, while style information has been effectively removed. In panel (b), style codes are projected onto a 2D space using t-SNE and colored by their corresponding style labels. Notably, clear style clusters emerge unsupervisedly, aligning with style labels.}}}
	\label{fig:latent_vis}
	 % \vspace{-1.5em}
\end{figure*}

%We make efforts to visualize the learned content code and style code for in-depth analysis.
\revise{Given that our content encoder accommodates motions of arbitrary length, we extract content codes from the~\cite{xia2015realtime} dataset. This dataset, unseen by our models, provides annotations for both style and content labels. Notably, the motions in this dataset are usually short, typically within 3s, which is insufficient to our style encoder. Therefore, for style codes, we takes the motions from dataset ~\citep{aberman2020unpaired} for visualization. The models are learned in unsupervised setting, using VAE as latent model.} 

\revise{\noindent\textbf{Content Code Visualization.} \Cref{fig:latent_vis} (a) visually presents 2D projections of our content codes. The content codes are colored by their content labels on the left and by their style labels on the right. To generate these projections, the temporal content codes are aggregated along the temporal dimension and then mapped to 2D space using t-SNE. When the content codes are colored by content label (e.g., \textit{walking, kicking}), distinct clusters aligned with the corresponding labels become apparent. However, when the same set of content codes is colored by their style label, these labels are evenly distributed within these clusters. This observation suggests that the content code adeptly captures the characteristics of various contents while effectively erasing style information.}

\revise{\noindent\textbf{Style Code Visualization.} \Cref{fig:latent_vis} (b) visualizes the style codes in a 2D space, color-coded by their style labels. Notably, these style labels were never used during model training. In contrast to the content code visualization in~\cref{fig:latent_vis} (a), the projected style codes exhibit a strong connection with the external style label annotations. This observation underscores the effectiveness of our style encoder in extracting style features from the motion corpus. }

\section{Loss Weight Analysis}
\label{sec:experiment}
\cref{tab:aberman_xia} presents more quantitative results of our models on ~\citep{aberman2020unpaired} and ~\citep{xia2015realtime} test sets. Specifically, we provide the ablation evaluations in both supervised (S) and unsupervised setting (U). For supervised setting, we conduct experiments on label-based stylization which also compares the diversity; and for unsupervised setting we adopt motion-based stylization. Note the base models are not necessarily our final models, here they are set only for reference.

\begin{table}[ht]
    % \centering
    \scalebox{0.76}{
    \begin{tabular}{l c c c c c c c c c c c c c}
    \toprule
    % \multirow{2}{*}{Methods} & \multicolumn{4}{c}{\DN (Coarse-grained)} & & \multicolumn{4}{c}{HumanAct(Fine-grained)} \\
    % \cline{2-5}
    % \cline{7-10}
    %                 & FID$\downarrow$ & Accuracy$\uparrow$ & Diversity$\rightarrow$& MModality$\rightarrow$ &  & FID$\downarrow$ & Accuracy$\uparrow$ & Diversity$\rightarrow$ & MModality$\rightarrow$\\
    \multirow{2}{*}{S / U}  & \multirow{2}{*}{$\lambda_{cyc}$} & \multirow{2}{*}{$\lambda_{kl}$}& \multirow{2}{*}{$\lambda_{hsa}$}&  \multicolumn{3}{c}{~\citep{aberman2020unpaired}} & ~ &\multicolumn{3}{c}{~\citep{xia2015realtime}} \\
    \cline{5-7}
    \cline{9-11}
    ~ & ~ & ~ & ~& Style Acc$\uparrow$ & Geo Dis$\downarrow$ & Div$\uparrow$& ~& Style Acc$\uparrow$ & Content Acc$\uparrow$ & Content FID$\downarrow$  \\
    \midrule
    % A & 0.1 & 0.1 & 1 & \et{0.942}{006} & \et{0.344}{003} & \et{0.050}{006} & & \et{0.933}{011} & \et{0.668}{014} & \et{0.193}{005} \\
    % A & 0.01 &  &  & \et{0.954}{007} & \et{0.357}{002} & \et{0.046}{005} & & \et{0.947}{007} & \et{0.625}{012} & \et{0.202}{004} \\
    % A & 1 &  &  & \et{0.960}{007} & \et{0.375}{003} & \et{0.060}{006} & & \et{0.959}{007} & \et{0.636}{015} & \et{0.209}{004} \\

    % A &  & 0.05 &  & \et{0.933}{006} & \et{0.348}{003} & \et{0.056}{005} & & \et{0.945}{007} & \et{0.726}{009} & \et{0.184}{006} \\
    % A &  &  & 0.1 & \et{0.942}{007} & \et{0.410}{003} & \et{0.153}{019} & & \et{0.920}{028} & \et{0.586}{014} & \et{0.205}{004} \\
    % A & 1 &  & 0.1 & \et{0.897}{010} & \et{0.433}{003} & \et{0.239}{024} & & \et{0.876}{011} & \et{0.642}{014} & \et{0.186}{005} \\
    S (base) &  0.1 & 0.01 & 0.1 & \et{0.937}{008} & \et{0.415}{003} & \et{0.153}{016} & & \et{0.913}{008} & \ets{0.669}{013} & \et{0.202}{006} \\

    \midrule
     &   & & 0.5 & \et{0.936}{008} & \ets{0.369}{003} & \et{0.091}{011} & & \et{0.924}{007} & \etb{0.706}{010} & \etb{0.197}{007} \\
    \cline{2-11}
    % A &   & 0 &  & \et{0.897}{006} & \et{0.429}{004} & \et{0.125}{016} & & \et{0.933}{009} & \et{0.619}{014} & \et{0.197}{005} \\
     &   & 0.001 &  & \etb{0.962}{006} & \et{0.429}{004} & \et{0.125}{016} & & \ets{0.933}{009} & \et{0.619}{014} & \ets{0.197}{005} \\
     
     &   & 0.1& & \et{0.940}{008} & \et{0.414}{004} & \et{0.141}{015} & & \et{0.914}{009} & \et{0.634}{013} & \et{0.209}{005} \\
    
    \cline{2-11}
    & 0.01 &  &  & \ets{0.955}{006} & \et{0.419}{003} & \et{0.107}{011} & & \etb{0.957}{007} & \et{0.609}{011} & \et{0.207}{006} \\
    &  1 & & & \et{0.880}{011} & \et{0.423}{003} & \ets{0.302}{026} & & \et{0.833}{011} & \et{0.625}{013} & \et{0.236}{006} \\
    \midrule
    U (base)& 1 & 0.01 & 0.1& \etb{0.804}{011} & \et{0.441}{003} & - & & \etb{0.814}{014} & \et{0.588}{010} & \et{0.217}{006} \\
    \midrule
    &  &  & 0.01& \ets{0.790}{015} & \et{0.489}{004} & - & & \et{0.761}{012} & \et{0.567}{016} & \et{0.224}{007} \\
    \cline{2-11}
    &  & 0.1 & & \et{0.659}{018} & \et{0.430}{004} & - & & \et{0.701}{014} & \et{0.619}{013} & \etb{0.190}{005} \\
    \cline{2-11}
    & 0.01 &  & & \et{0.669}{013} & \etb{0.388}{003} & - & & \et{0.671}{015} & \etb{0.641}{012} & \ets{0.206}{006} \\

    & 0.1 &  & & \et{0.739}{015} & \ets{0.420}{004} & - & & \ets{0.762}{016} & \ets{0.619}{014} & \et{0.214}{007} \\

%    & 0.1 & 0.1 & 1& \et{0.325}{014} & \etb{0.281}{002} & - & & \et{0.426}{015} & \etb{0.660}{011} & \ets{0.205}{007} \\

    \bottomrule
    \end{tabular}
    }
    \caption{\footnotesize{Effect of hyper-parameters of \textit{ours (A)} on the ~\citep{aberman2020unpaired} and ~\citep{xia2015realtime} test sets. $\pm$ indicates 95\% confidence interval. \textbf{Bold} face indicates the best result, while \underline{underscore} refers to the second best. (S) and (U) denote \textit{supervised} and \textit{unsupervised} setting. For (S), we present results of label-based stylization; and for (U), we present motion-based stylization.}}
    \label{tab:aberman_xia}
    	 % \vspace{-1em}

\end{table}

\paragraph{Effect of $\lambda_{hsa}$.} Homo-style alignment ensures the style space of the sub-clips from one motion sequence to be close to each other; it is an important self-supervised signal in our approach. Increasing the weight of homo-style commonly helps style modeling (style accuracy) and content preservation (content accuracy, FID), which however also comes with lower diversity. A common observation is that the performance on style and content always contradicts with the diversity. It could be possibly attributed to the inherently limited diversity in our training dataset~\citep{aberman2020unpaired}, which is collected by one person performing several styles.

\paragraph{Effect of $\lambda_{kl}$.} $\lambda_{kl}$ weighs how much the overall style space aligns with the prior distribution $\mathcal{N}(\mathbf{0}, \mathbf{I})$. Smaller $\lambda_{kl}$ usually increases the capacity of the model exploiting styles, which on the other hand deteriorate the performance on content maintenance and diversity.

\paragraph{Effect of $\lambda_{cyc}$.} Cycle reconstruction constraint plays an important role in unsupervised setting. In supervised setting, strong cycle reconstruction constraint is detrimental to style modeling. In contrast, while learning unsupervisedly, strengthening the cycle constraint enhances the performance on style transferring, and at the same time compromises the preservation of content.

\begin{table}[ht]
    % \centering
    \scalebox{0.73}{
    \begin{tabular}{c c c c c c c c c c c c}
    \toprule
    % \multirow{2}{*}{Methods} & \multicolumn{4}{c}{\DN (Coarse-grained)} & & \multicolumn{4}{c}{HumanAct(Fine-grained)} \\
    % \cline{2-5}
    % \cline{7-10}
    %                 & FID$\downarrow$ & Accuracy$\uparrow$ & Diversity$\rightarrow$& MModality$\rightarrow$ &  & FID$\downarrow$ & Accuracy$\uparrow$ & Diversity$\rightarrow$ & MModality$\rightarrow$\\
     \multirow{2}{*}{$\lambda_{l1}$} & \multirow{2}{*}{$\lambda_{sms}$}&  \multicolumn{3}{c}{~\citep{aberman2020unpaired}} & ~ &\multicolumn{3}{c}{~\citep{xia2015realtime}} \\
    \cline{3-5}
    \cline{7-10}
     ~ & ~ & MPJPE (Recon)$\downarrow$ &Style Acc$\uparrow$ & Style FID$\downarrow$& ~& MPJPE (Recon)$\downarrow$&Style Acc$\uparrow$ & Content Acc$\uparrow$ & Content FID$\downarrow$  \\
    \midrule
        % - & Ours (V) \\
      0.001 & 0.001 & \textbf{39.4} & \etb{0.945}{007} & \etb{0.020}{002} & & \textbf{62.5}&\etb{0.926}{008} & \etb{0.674}{011} & \etb{0.189}{005} \\
      \midrule
      0.1 & 0.1 & 360.1 & \et{0.862}{010} & \et{0.041}{004} & & 431.8&\et{0.804}{011} & \et{0.589}{012} & \et{0.276}{007} \\
      0.01 & 0.01 & 180.4 & \et{0.873}{010} & \et{0.041}{004} & & 250.5&\et{0.830}{009} & \et{0.656}{012} & \et{0.244}{007} \\
      0.0001 & 0.0001 & 77.6 & \et{0.857}{010} & \et{0.042}{003} & & 130.9&\et{0.901}{011} & \et{0.661}{013} & \et{0.239}{007} \\

    \bottomrule
    \end{tabular}
    }
    \caption{\revise{\footnotesize{Effect of hyper-parameters of autoencoder on the ~\citep{aberman2020unpaired} and ~\citep{xia2015realtime} test sets. $\pm$ indicates 95\% confidence interval. \textbf{Bold} face indicates the best result, while \underline{underscore} refers to the second best. Results of motion-based stylization in supervised setting are presented. MPJPE is measured in millimeter.}}}
    \label{tab:ae_effect}
    	 % \vspace{-1em}

\end{table}

\noindent\revise{\textbf{Effect of Autoencoder Hyper-Parameters.} In~\cref{tab:ae_effect}, we investigate the impact of autoencoder hyper-parameters ($\lambda_{l1}$ and $\lambda_{sms}$) on both motion reconstruction and stylization performance. Specifically, $\lambda_{l1}$ encourages sparsity in latent features, while $\lambda_{sms}$ enforces the smoothness of temporal features. Through experimentation, we identify an optimal set of hyper-parameters with $\lambda_{l1}=0.001$ and $\lambda_{sms}=0.001$, which yields optimal performance in both reconstruction and stylization tasks. Notably, imposing excessive penalties on smoothness and sparsity proves detrimental to the model's capabilities, resulting in lower reconstruction quality. Additionally, we observe a substantial correlation between reconstruction and stylization performance, indicating that better reconstruction often translates to improved stylization.}

% \begin{table}[t]
%     \centering
%     \scalebox{0.76}{
%     \begin{tabular}{l c c c c }
%     \toprule
%     % \multirow{2}{*}{Methods} & \multicolumn{4}{c}{\DN (Coarse-grained)} & & \multicolumn{4}{c}{HumanAct(Fine-grained)} \\
%     % \cline{2-5}
%     % \cline{7-10}
%     %                 & FID$\downarrow$ & Accuracy$\uparrow$ & Diversity$\rightarrow$& MModality$\rightarrow$ &  & FID$\downarrow$ & Accuracy$\uparrow$ & Diversity$\rightarrow$ & MModality$\rightarrow$\\
%     Model & Style Acc$\uparrow$ & Style FID$\downarrow$ & Geo Dis$\downarrow$& Foot Skating$\downarrow$  \\
%     \midrule
%     Ours (S) & \etb{0.945}{007} & \etb{0.020}{002} & \et{0.344}{002} & \etb{0.130}{001} \\
%     Ours \textit{w/o} GMP (S) & \et{0.942}{006} & \et{0.021}{002} & \etb{0.341}{003} & \et{0.141}{001} \\
%     \midrule
%     Ours (U) & \etb{0.840}{010} & \etb{0.036}{003} & \etb{0.478}{004} & \etb{0.102}{001} \\
%     Ours \textit{w/o} GMP (U) & \et{0.817}{013} & \et{0.038}{003} & \et{0.478}{003} & \et{0.116}{001} \\
%     \bottomrule
%     \end{tabular}
%     }
%     \caption{\footnotesize{Effect of global motion prediction (GMP, ~\cref{subsec:gmp}) on the \citep{aberman2020unpaired} test set. $\pm$ indicates 95\% confidence interval. \textbf{Bold} face indicates the best result. (S) and (U) denote \textit{supervised} and \textit{unsupervised} setting. Results of motion-based stylization is presented.}}
%     \label{tab:gmp}
%     	 % \vspace{-1em}
% \end{table}

\section{Implementation Details}
\label{sec:implementation}

% \paragraph{Implementation Details.} 
Our models are implemented by Pytorch. Motion encoder $\mathcal{E}$ and decoder $\mathcal{D}$ consists of 2 1-D convolution layers; global motion regressor is a 3-layer 1D convolution network. The content encoder $\mathrm{E}_c$ and style encoder $\mathrm{E}_s$ are also downsampling convolutional networks, where style encoder contains a average pooling layer before the output dense layer. The spatial dimensions of content and style code are both 512. Detailed model architecture is provided in~\cref{fig:fullarc,fig:mvae}. The values of $\lambda_{kld}^l$, $\lambda_{l1}$ and $\lambda_{sms}$ are all set to 0.001, and dimension $D_z$ of $\mathbf{z}$ is 512. During training our latent stylization network, the value of $\lambda_{hsa}$, $\lambda_{cyc}$ and $\lambda_{kl}$ are (1, 0.1, 0.1) and (0.1, 1, 0.01) in supervised setting and unsupervised setting, respectively. 

\subsection{Model Structure}
\begin{figure*}[ht]
	\centering
	\includegraphics[width=0.8\linewidth]{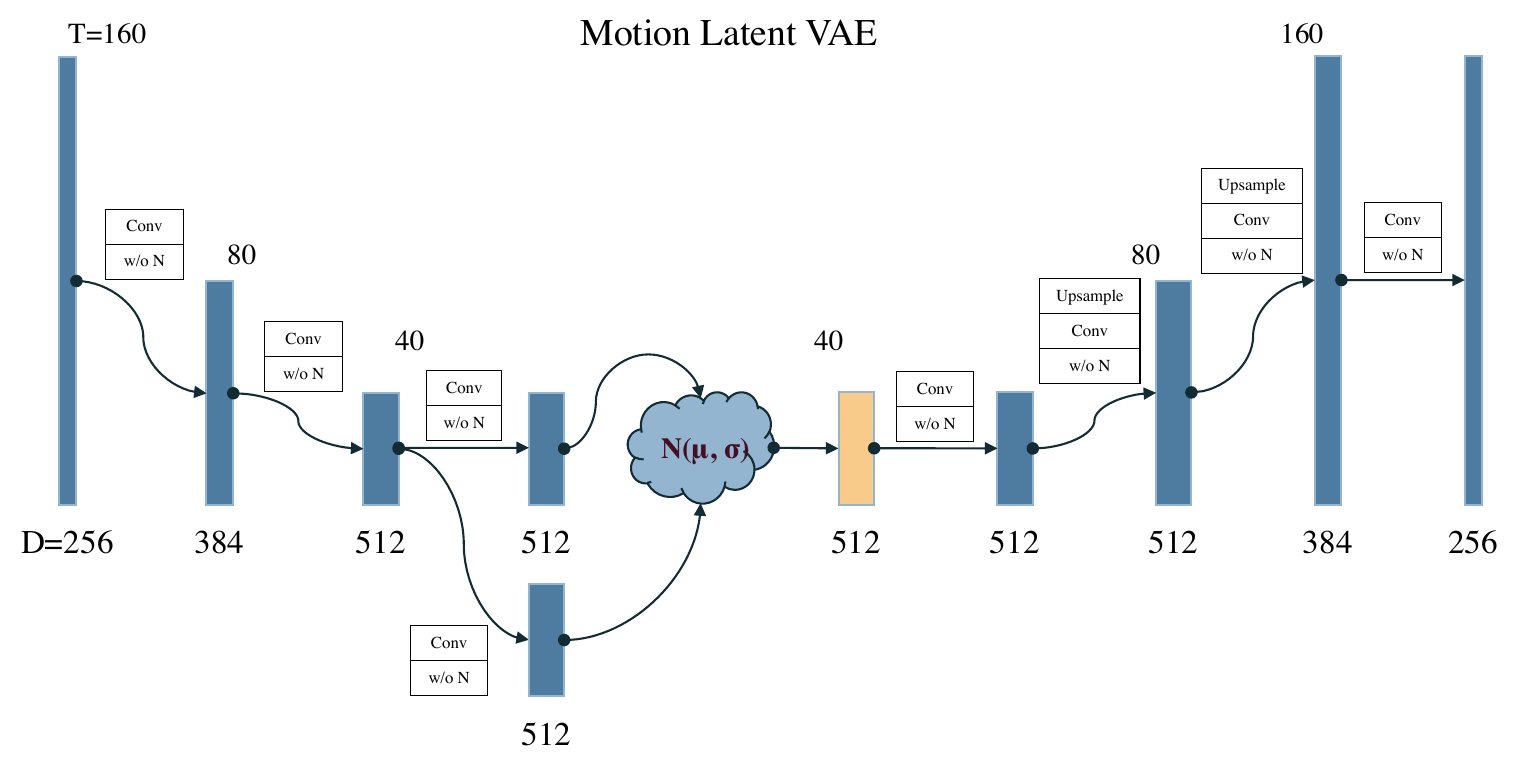}
 	 \vspace{-1.5em}
	\caption{Detailed architecture of our VAE based motion latent model. The AE based latent model keeps only one convolution branch before the latent space. All convolutions, except the last layer of encoder, decoder and generator, use kernel size of 3.}
	\label{fig:mvae}
	 \vspace{-1.5em}
\end{figure*}

\begin{figure*}[ht]
	\centering
	\includegraphics[width=\linewidth]{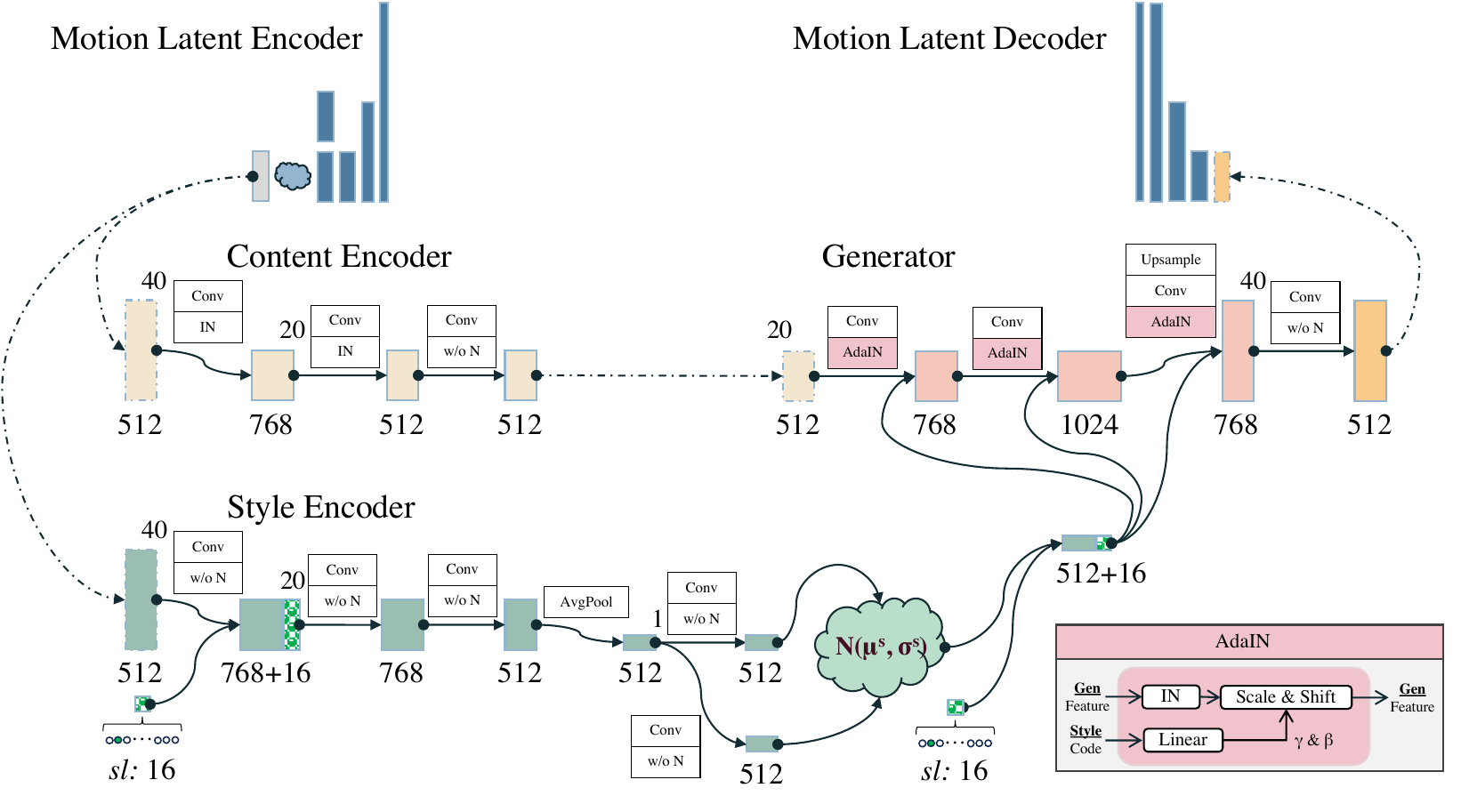}
 	 \vspace{-1.5em}
	\caption{Detailed architecture of our motion latent stylization model in supervised setting. In unsupervised setting, the style label input is dropped. All convolutions, except the last layer of encoders and generator, use kernel size of 3.}
	\label{fig:fullarc}
	 \vspace{-1.5em}
\end{figure*}

% \section{Architecture}
\label{sec:architecture}
The detailed architectures of our motion latent auto-encoder and motion latent stylization model are illustrated in \Cref{fig:mvae} and \Cref{fig:fullarc} respectively, where "w/o N", "IN" and "AdaIN" refer to without-Normalization, Instance Normalization and Adaptive Instance Normalization operations~\citep{huang2017arbitrary}. Dropout and Activation layer are omitted for simplicity. %All convolutions, except the last layer of encoder and decoder, use kernel size of 3.

\subsection{Data Processing}
We mostly adopt the pose processing procedure in~\citep{guo2022generating}. In short, a single pose is represented by a tuple of root angular velocity, root linear velocity, root height, local joint positions, velocities, 6D rotations~\citep{zhou2019continuity} and foot contact labels, resulting in 260-D pose representation. Meanwhile, all data is downsampled to 30 FPS, augmented by mirroring, and applied with Z-nomalization.

\subsection{Baseline Implementation}
\label{subsec:baseline}
For a fair comparison, we adapt the baseline models with minimal changes from their official implementations, training them on the same data splits. More specifically, without violating their design of input representation and networks, all the re-implemented baseline methods strictly load the same preprocessed data for training.

\textbf{\citep{aberman2020unpaired}.} Due to the intentional dual representations for style and content inputs in \citep{aberman2020unpaired}, we make some modifications in the dataloader. We first recover the raw 21-joints structural motion data from the preprocessed data, and convert them into 84-D rotation-based content feature along with 4-D global motion, and 63-D position-based style feature, using their motion parsing function. In addition, we  modified the channels of network input/output layers to fit the adapted data. Since our experiments solely consider style from 3-D motions, we disable the 2-D branch as well as the related loss functions. However, we suffer from extremely unstable training process and poor results using the same hyper-parameters. It may result from the length extension of motion sequence (now 160 vs original 32) and inherent flaws of GANs~\citep{zhu2017unpaired, karras2019style}. Thus, empirically, we lower the coefficient for adversarial loss $\alpha_{adv}$ from 1 to 0.5, and update the frequency of discriminator training from 1 per-iteration to 0.2 per-iteration.

\begin{figure*}[thb]
	\centering
	\includegraphics[width=0.9\linewidth]{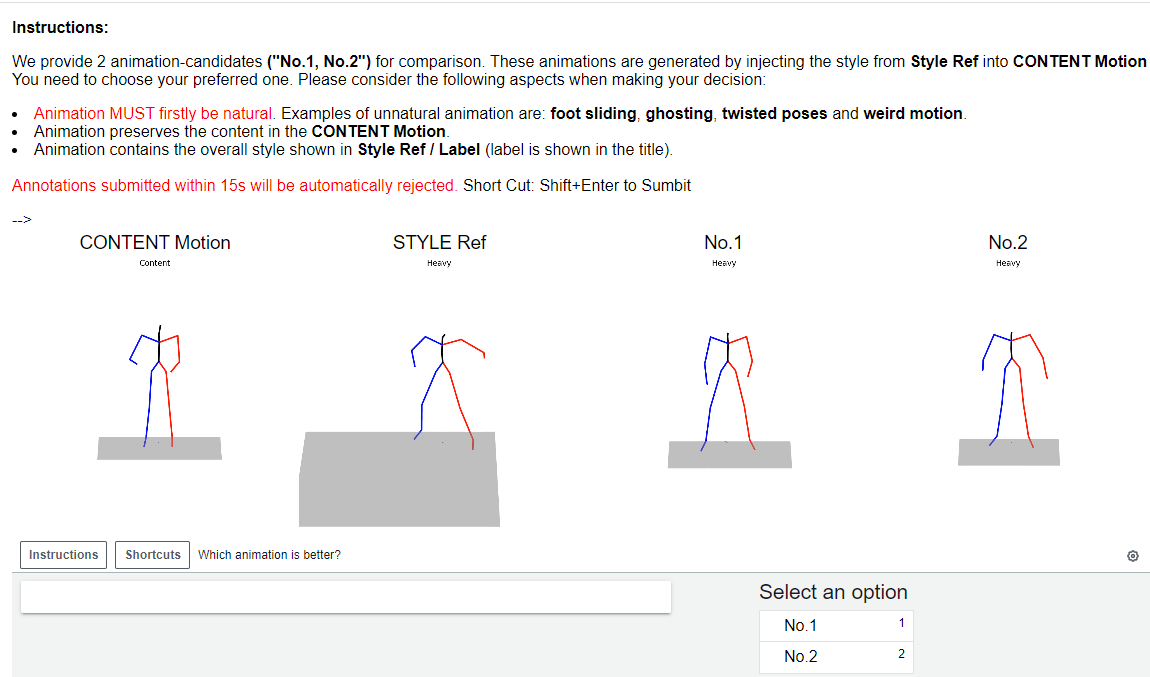}
 	 % \vspace{-1.5em}
	\caption{User study interface on Amazon Mechanical Turk.}
	\label{fig:userstd}
	 % \vspace{-1.5em}
\end{figure*}

\textbf{\citep{park2021diverse}.} We extract 63-D joint position feature and 126-D joint rotation feature from our preprocessed data, catering for the designated dataloader in \citep{park2021diverse}. Meanwhile, we replace the original 4-D quaternion with the equally functional 6-D rotation \citep{zhou2019continuity} without any loss of capability. Their model design is limited to fixed motion length, due to the un-scalable linear layer. Therefore, during evaluation on ~\citep{xia2015realtime} test set, we duplicate the sequence to meet the 160-length setting and then extract the corresponding result from the output.

\textbf{\citep{jang2022motion}} takes the motion representation building from per-joint's 6-D position-based feature (i.e. position, velocity) and 6-D rotation-based feature (i.e. upward direction, forward direction) which is almost coherent with our preprocessed data. Thus, we directly re-organize our data to serve the baseline \citep{jang2022motion}, keeping everything else unchanged.

\section{Evaluation Metric}
\label{sec:metric}
\textbf{Why certain metrics are not used across all datasets?} Given our latent stylization models are trained on ~\citep{aberman2020unpaired}, CMU Mocap~\citep{cmu2021mocap} and ~\citep{xia2015realtime} aims to emphasize zero-shot performance on Style Precision and Content Preservation respectively. Style classifier is trained on ~\citep{aberman2020unpaired}, where all style motions come from. Compared to ~\citep{aberman2020unpaired} and CMU Mocap, ~\citep{xia2015realtime} is quite small (570 clips), comprising variable-length short motions ($<$ 3s). Style FID isn't computed for ~\citep{xia2015realtime} due to substantial length differences between the style motion (from \citep{aberman2020unpaired}, 5.3s) and output motion ($<$ 3s). Content classifier is trained only on ~\citep{xia2015realtime} to evaluate the Content Preservation as content labels are only available on this dataset. Since there is no evidence that this content classifier can generalize to other datasets, we only use it for ~\citep{xia2015realtime}.

\section{User Study}
\label{sec:user_study}
The interface of the user study on Amazon Mechanical Turk for our experiments is shown in \Cref{fig:userstd}. Since motion style is not as obvious as other qualitative attributes for common users, to simplify the study, we only compare one baseline result with ours each time. Moreover, for introduction, we briefly explain the concept of motion stylization, presenting the content motion as well as style motion for reference. Users are instructed to choose their preferred results over two generated stylization results based on judgement on \textit{naturalism}, \textit{content preservation} and \textit{style visibility}. This study only involves users that are recognized as \textbf{master} by AMT.

\section{Interpolation}
\label{sec:interpolation}
\begin{figure}[h]
\begin{minipage}{0.6\textwidth}
\includegraphics[width=\linewidth]{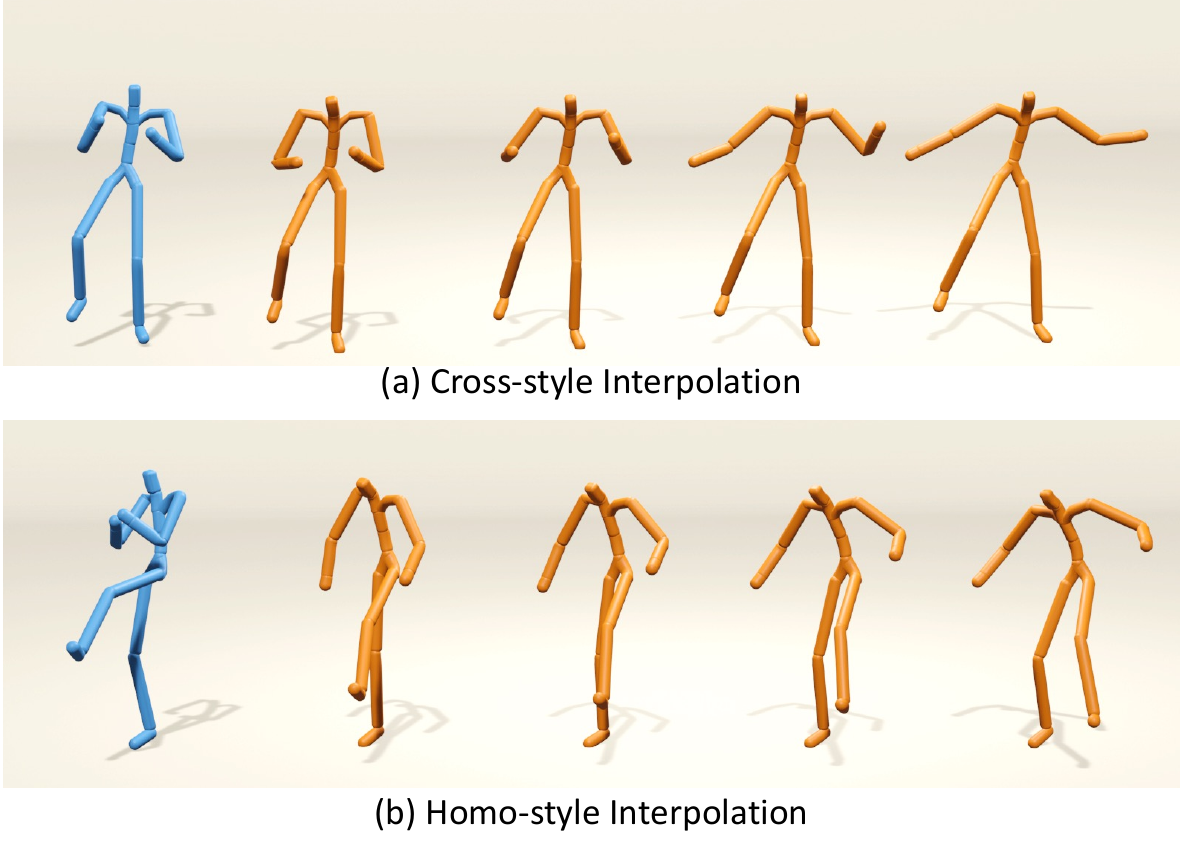}
\end{minipage}%
\hspace{0.03\textwidth}
\begin{minipage}{0.36\textwidth}
\caption{\footnotesize{\textbf{Style Interpolation.} \textbf{(a)} Cross-style interpolation in unsupervisedly learned style space. Styles are interpolated between style codes of \textit{sneaky} (left) and \textit{heavy} (right) motions. \textbf{(b)} Homo-style interpolation in supervisedly learned style space. With style label \textit{heavy} as condition input, styles are interpolated between two style codes that randomly sampled from $\mathcal{N}(\mathbf{0}, \mathbf{I})$. One key pose for each motion is displayed.}}
\label{fig:interpolate}
\end{minipage}
\end{figure}

We present the results of interpolation in the respective style spaces learned unsupervisedly \cref{fig:interpolate}(a) and supervisedly \cref{fig:interpolate}(b). We are able to interpolate between styles from different labels in unsupervised setting. Specifically, two style codes are extracted from \textit{sneaky} motion and \textit{heavy} motion respectively. Then we mix these two style codes through linear interpolation, and apply them to stylize the given content motion. In supervised setting, the generator is conditioned on a specific style label. Here, we interpolate styles between two random style codes sampled from the prior distribution $\mathcal{N}(\mathbf{0}, \mathbf{I})$. Stylization results are produced conditioned a common style label, \textit{heavy}. From ~\Cref{fig:interpolate}, we can observe the smooth transitions along the interpolation trajectory of two different style codes. Please refer to our supplementary video for better visualization.

% \begin{figure}[hbt]
% \captionsetup{justification=raggedleft, singlelinecheck=false}
% % \centering
% \includegraphics[width=0.6\textwidth]{figures/interpolation.pdf}
% \caption{This is a caption on the right side.}
% \end{figure}

\section{Limitations and Failure Cases}
\label{sec:failure_cases}

\begin{figure}[thb]
\begin{minipage}{0.7\textwidth}
\includegraphics[width=\linewidth]{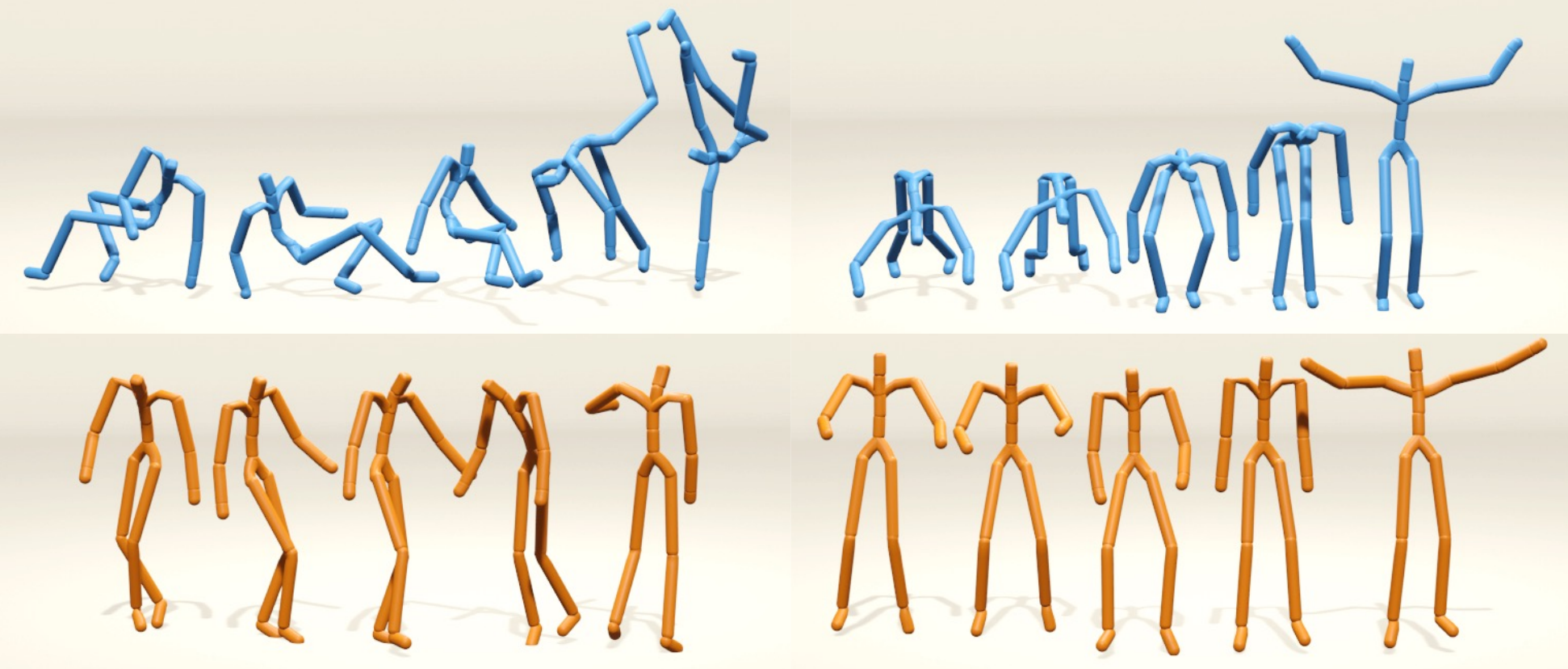}
\end{minipage}%
\hspace{0.01\textwidth}
\begin{minipage}{0.27\textwidth}
\caption{Failure cases. Top row shows content motion; bottom row shows our corresponding results. Stylization results of breaking dance motion (left) and push-up motion (right) using \textit{happy} style label are displayed.}
\label{fig:fail}
\end{minipage}
\end{figure}

Firstly, our model may encounter difficulties when the input motion substantially deviates from our training data. \Cref{fig:fail} presents two failed stylization results on rare content actions, \textit{i.e.,} breaking dance and push-up. Given that our model has only seen standing motions during training, it commonly fails to reserve the lower-body movements in these two cases. Interestingly, our model can still retain the general motions of upper-body. 

Secondly, the underlying reason for different performance of ours(V) and ours(A) on for example, diversity, style and content accuracy, remains unclear. 

Lastly, certain styles are inherently linked to specific content characteristics, particularly within the datasets of \citep{aberman2020unpaired,xia2015realtime}. For instance, styles like old, depressed and lazy typically relate to slow motions, while happy, hurried, angry motions tend to be fast. As our stylization process aims to preserve content information, including speed, there could be contradictions with these style attributes. For instance, stylizing an slow motion with a hurried style might not yield an outcome resembling a hurried motion. We acknowledge this aspect for potential exploration in future studies.
%As a future step, we will include motion data with more diversity and coverage to tackle these limitations. %may large-scale motion data, conditions, such as content global motion correspondence, and large-scale motion data to tackle these limitations.
% \section{Long Motion Stylization}

\end{document}